\documentclass[journal]{IEEEtran}

\usepackage{booktabs} 
\usepackage{algorithm}
\usepackage{algpseudocode}
\usepackage{amssymb}
\usepackage{amsmath}
\usepackage{upgreek}
\usepackage{pdflscape}
\usepackage{afterpage}
\usepackage{rotating}
\usepackage{graphicx}
\usepackage{svg}
\usepackage{ragged2e}
\usepackage{enumitem}
\usepackage[hidelinks]{hyperref}
\usepackage{cite}
\usepackage{caption}
\usepackage{multirow}

\DeclareMathOperator*{\argmax}{arg\,max}

\DeclareMathOperator*{\maximize}{maximize}

\begin{document}

\title{Deep Reinforcement Learning for\\ Sequence-to-Sequence Models}

\author{Yaser Keneshloo, Tian Shi, Naren Ramakrishnan, Chandan K. Reddy,~\IEEEmembership{Senior Member,~IEEE}
                \thanks{Y. Keneshloo, T. Shi, N. Ramakrishnan, and C. K. Reddy are with the Discovery Analytics Center, Department of Computer Science at Virginia Tech, Arlington, VA. \{yaserkl,tshi\}@vt.edu, \{naren,reddy\}@cs.vt.edu. Corresponding author: yaserkl@vt.edu.}
		\thanks{This paper is currently under review in IEEE Transactions on Neural Networks and Learning Systems}
}

%

\maketitle

\begin{abstract}
In recent times, sequence-to-sequence (seq2seq) models have gained a lot of popularity and provide state-of-the-art performance in a wide variety of tasks such as machine translation, headline generation, text summarization, speech to text conversion, and image caption generation.
The underlying framework for all these models is usually a deep neural network comprising an encoder and a decoder.
Although simple encoder-decoder models produce competitive results, many researchers have proposed additional improvements over these seq2seq models, e.g., using an attention-based model over the input, pointer-generation models, and self-attention models.
However, such seq2seq models suffer from two common problems: 1) \textit{exposure bias} and 2) \textit{inconsistency between train/test measurement}.
Recently, a completely novel point of view has emerged in addressing these two problems in seq2seq models, leveraging methods from reinforcement learning (RL).
In this survey, we consider seq2seq problems from the RL point of view and  provide a formulation combining the power of RL methods in decision-making with seq2seq models that enable remembering long-term memories.
We present some of the most recent frameworks that combine concepts from RL and deep neural networks.
Our work aims to provide insights into some of the problems that inherently arise with current approaches and how we can address them with better RL models.
We also provide the source code for implementing most of the RL models discussed in this paper to support the complex task of abstractive text summarization and provide some targeted experiments for these RL models, both in terms of performance and training time.
\end{abstract}

\IEEEpeerreviewmaketitle

\section{Introduction}\label{intro}
\IEEEPARstart{S}{equence}-to-sequence (seq2seq) models constitute a common framework for solving sequential problems~\cite{sutskever2014sequence}. In seq2seq models, the input is a sequence of certain data units and the output is also a sequence of data units. Traditionally, these models are trained using a ground-truth sequence via a mechanism known as \textit{teacher forcing}~\cite{bengio2015scheduled}, where the teacher is the ground-truth sequence.
However, due to some of the drawbacks of this training approach, there has been significant line of research connecting inference of these models with reinforcement learning (RL) techniques. In this paper, we aim to summarize such research in seq2seq training utilizing RL methods to enhance the performance of these models and discuss various challenges that arise when applying RL methods to train a seq2seq model. We intend for this paper to provide a broad overview on the strength and complexity of combining seq2seq training with RL training and to guide researchers in choosing the right RL algorithm for solving their problem. In this section, we will briefly introduce the working of a simple seq2seq model and outline some of the problems that are inherent to seq2seq models. We will then provide an introduction to RL models and explain how these models could solve the problems of seq2seq models.

\subsection{Seq2seq Framework}
\label{section:introsec}
Seq2seq models are common in various applications ranging from machine translation~\cite{luong2015effective,wu2016google,vaswani2018tensor2tensor,bahdanau2014neural,cho2014learning,shen2015minimum}
, news headline generation~\cite{rush2015neural,chopra2016abstractive}
, text summarization~\cite{nallapati2016abstractive,see2017get,paulus2017deep,nallapati2017summarunner}
, speech-to-text applications~\cite{graves2013speech,bahdanau2016end,amodei2016deep,arik2017deep}
, and image captioning~\cite{xu2015show,vinyals2015show,karpathy2015deep}.

\begin{figure}
\centering
\includegraphics[scale=0.165]{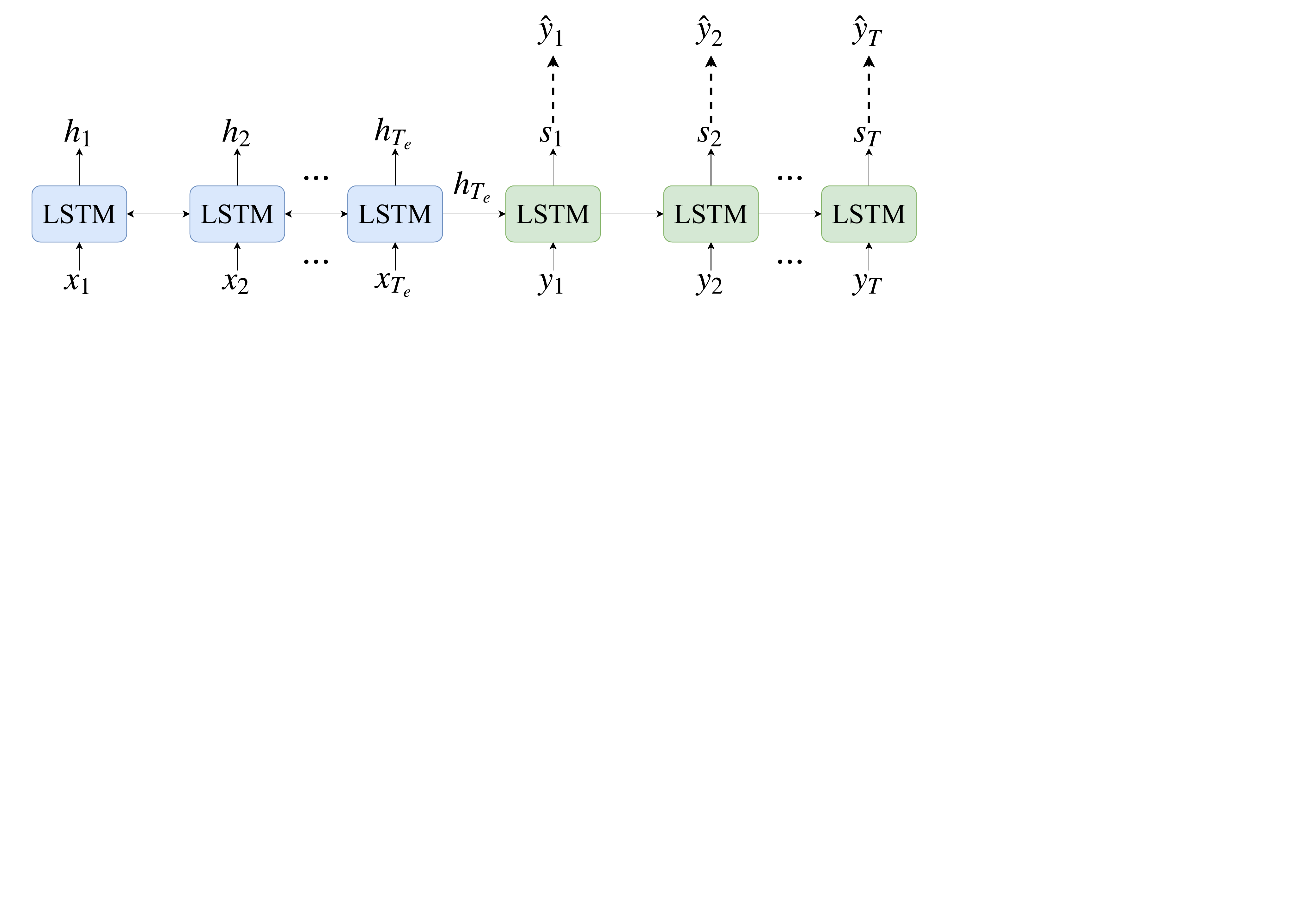}
\caption{A simple seq2seq model. The blue boxes correspond to the encoder part which has $T_e$ units.
The green boxes correspond to the decoder part which has $T$ units.
}
\label{fig:seq2seq}
\end{figure}

In recent years, the general framework for solving these problems uses deep neural networks that comprise two main components: an encoder which reads the sequence of input data and a decoder which uses the output generated by the encoder to produce the sequence of final outputs.
Fig~\ref{fig:seq2seq} gives a schematic of this simple yet effective framework.
The encoder and decoder are usually implemented by recurrent neural networks (RNN) such as Long Short-Term Memory (LSTM)~\cite{hochreiter1997long}.
The encoder takes a sequence of length $T_e$ inputs\footnote{In this paper, we use input/output and action interchangeably since choosing the next input is akin to choosing the next action and generating the next output is akin to generating the next action.}, $X=\{x_1, x_2, \cdots, x_{T_e}\}$, where $x_t\in \mathcal{A}=\{1,\cdots, |\mathcal{A}|\}$ is a single input coming from a range of possible inputs ($\mathcal{A}$), and generates the output state $h_t$.
In addition, each encoder receives the the previous encoder's hidden state, $h_{t-1}$, and if the encoder is a bidirectional LSTM, it will also receive the state from the next encoder's hidden state, $h_{t+1}$, to generate its current hidden state $h_t$.
The decoder, on the other hand, takes the last state from the encoder, i.e., $h_{T_e}$ and starts generating an output of size $T<T_e$, $\hat{Y}=\{\hat{y}_1,\hat{y}_2,\cdots,\hat{y}_T\}$, based on the current state of the decoder $s_t$ and the ground-truth output $y_t$. The decoder could also take as input an additional context vector $c_t$, which encodes the context to be used while generating the output~\cite{rush2015neural}. The RNN learns a recursive function to compute $s_t$ and outputs the distribution over the next output:
\begin{equation}
\begin{array}{l}
h_{t^{'}} =\Phi_\theta(x_{t^{'}},h_{t})\\
s_{t^{'}} =\Phi_\theta(y_t,s_t/h_{T_e},c_t)\\
\hat{y}_{t^{'}} \sim \pi_{\theta}(y|\hat{y}_t, s_{t^{'}})
\label{eq:seq2seqp}
\end{array}
\end{equation}
where $t^{'}= t+1$, $\theta$ denotes the parameters of the model, and the function for $\pi_{\theta}$ and $\Phi_\theta$ depends on the type of RNN. A simple Elman RNN~\cite{elman1990finding} would use a sigmoid function for $\Phi$ and a softmax function for $\pi$~\cite{sutskever2014sequence}:
\begin{equation}
\begin{array}{l}
s_{t^\prime} = \sigma(W_1 y_{t} + W_2 s_{t} + W_3 c_{t})\\
o_{t^\prime} = \textrm{softmax}(W_4 s_{t^\prime} + W_5 c_{t})
\end{array}
\end{equation}
\noindent
where $o_t$ is the output distribution of size $|\mathcal{A}|$ and the output $\hat{y}_t$ is selected from this distribution.
$W_1$, $W_2$, $W_3$, $W_4$, and $W_5$ are matrices of learnable parameters of sizes $W_{1,2,3} \in R^{d\times d}$ and $W_{4,5} \in R^{d\times |\mathcal{A}|}$, where $d$ is the size of the input representation (e.g., size of the word embedding in text summarization).
The input to the first decoder is a special input indicating the beginning of a sequence, denoted by $y_0 = \emptyset$ and the first forward hidden state $h_0$ and the last backward hidden state $h_{T_e+1}$ for the encoder are set to a zero vector.
Moreover, the first hidden state for decoder $s_0$ is set to the output that is received from the last encoding state, i.e., $h_{T_e}$.

The most widely used method to train the decoder for sequence generation is called the teacher forcing algorithm~\cite{bengio2015scheduled}, which minimizes the maximum-likelihood loss at each decoding step.
Let us define $y = \{y_{1}, y_{2}, \cdots , y_{T}\}$ as the ground-truth output sequence for a given input sequence $X$. The maximum-likelihood training objective is the minimization of the following cross-entropy (CE) loss:
\begin{equation}
\mathcal{L}_{CE}=-\sum_{t=1}^{T}\log{\pi_{\theta}(y_{t}|y_{t-1},s_t, c_{t-1},X)}
\label{eq:cel}
\end{equation}
Once the model is trained with the above objective, the model generates an entire sequence as follows:
Let $\hat{y}_t$ denotes the action (output) taken by the model at time $t$.
Then, the next action is generated by:
\begin{equation}
\hat{y}_{t^{'}} = \argmax_{y} \pi_{\theta}(y|\hat{y}_t, s_{t^{'}})
\label{eq:inf}
\end{equation}
This process could be improved by using beam search to find a reasonable good output sequence~\cite{cho2014learning}. Now, given the ground-truth output $Y$ and the model generated output $\hat{Y}$, the performance of the model is evaluated with a specific measure. In seq2seq problems, discrete measures such as $\textrm{ROUGE}$~\cite{lin2004rouge}, $\textrm{BLEU}$~\cite{papineni2002bleu}, $\textrm{METEOR}$~\cite{banerjee2005meteor}, and $\textrm{CIDEr}$~\cite{vedantam2015cider} are used to evaluate the model. For instance, $\textrm{ROUGE}_l$, an evaluation measure for textual seq2seq tasks, uses the largest common substring between $Y$ and $\hat{Y}$ to evaluate the goodness of the generated output. Algorithm~\ref{alg:seq2seq} shows these steps.

\begin{algorithm}[t]
\begin{algorithmic}
\footnotesize
\State \textbf{Input}: Input sequences ($X$) and ground-truth output sequences ($Y$).
\State \textbf{Output}: Trained seq2seq model.
\State \textbf{Training Steps}:
\For{batch of input and output sequences $X$ and $Y$}
    \State Run encoding on $X$ and get the last encoder state $h_{T_e}$.
    \State Run decoding by feeding $h_{T_e}$ to the first decoder and obtain the
    \State sampled output sequence $\hat{Y}$.
    \State Calculate the loss according to Eq. (\ref{eq:cel}) and update the parameters
    \State of the model.
\EndFor
\State \textbf{Testing Steps}:
\For{batch of input and output sequences $X$ and $Y$}
	\State Use the trained model and Eq. (\ref{eq:inf}) to sample the output $\hat{Y}$
	\State Evaluate the model using a performance measure, e.g., $\textrm{ROUGE}$
\EndFor
\end{algorithmic}
\caption{Training a simple seq2seq model}
\label{alg:seq2seq}
\end{algorithm}

\subsection{Problems with Seq2seq Models}
\label{section:seq2seqproblem}
One of the main issues with the current seq2seq models is that minimizing $\mathcal{L}_{CE}$ does not always produce the best results for the above discrete evaluation measures.
Therefore, using cross-entropy loss for training a seq2seq model creates a mismatch in generating the next action during training and testing.
As shown in Fig~\ref{fig:seq2seq} and also according to Eq. (\ref{eq:cel}), during training, the decoder uses the two inputs, the previous output state $s_{t-1}$ and the ground-truth input $y_t$, to calculate its current output state $s_t$ and uses it to generate the next action, i.e., $\hat{y}_t$.
However, at the test time, as given in Eq. (\ref{eq:inf}), the decoder completely relies on the previously generated action from the model distribution to predict the next action, since the ground-truth data is not available anymore.
Therefore, in summary, the input to the decoder is from the ground-truth during training, but the input comes from the model distribution during model testing.
This \textit{exposure bias}~\cite{ranzato2015sequence} results in error accumulation during the output generation at test time, since the model has never been exclusively exposed to its own predictions during training.
To avoid the \textit{exposure bias} problem, we need to remove the ground-truth dependency during training and use only the model distribution to minimize Eq. (\ref{eq:cel}).
One way to handle this situation is through the scheduled sampling method~\cite{bengio2015scheduled} or Gibbs sampling~\cite{su2018incorporating}.
In scheduled sampling, the model is first pre-trained using cross-entropy loss and will subsequently and slowly replace the ground-truth with a sampled action from the model.
Therefore, a decision is randomly taken to whether use the ground-truth action with probability $\epsilon$, or an action coming from the model itself with probability $(1-\epsilon)$.
When $\epsilon = 1$, the model is trained using Eq. (\ref{eq:cel}), and when $\epsilon = 0$ the model is trained based on the following loss:

\begin{equation}
\mathcal{L}_{\textrm{Inference}}=-\sum_{t=1}^{T}\log{\pi_{\theta}(\hat{y}_{t}|\hat{y}_{1}, \cdots, \hat{y}_{t-1},s_t, c_{t-1},X)}
\label{eq:infloss}
\end{equation}

Note the difference between this equation and CE loss; in CE the ground-truth output $y_t$ is used to calculate the loss, while in  Eq. (\ref{eq:infloss}), the output of the model $\hat{y}_t$ is used to calculate the loss.

Although scheduled sampling is a simple way to avoid the exposure bias, due to its random selection between choosing ground-truth output or model output, it does not provide a clear solution for the back-propagation of error and therefore it is statistically inconsistent~\cite{huszar2015not}.
Recently, Goyal \textit{et al.}~\cite{goyal2017differentiable} proposed a solution for this problem by creating a continuous relaxation over the argmax operation to create a differentiable approximation of the greedy search during the decoding steps.

As yet another line of research on avoiding the exposure bias problem, adversarial generative models are also proposed for various seq2seq models~\cite{yu2017seqgan,lin2017adversarial,guo2017long,che2017maximum}.
In general, adversarial models are comprised of a discriminator and a generator~\cite{goodfellow2014generative}.
The generator tries to generate data similar to the ground-truth data while the discriminator's job is to discern whether the generated data is close to real data or it is a fake.
Finally, the generator takes the feedback from the discriminator and optimizes its actions towards generating higher quality data.
Since generator will only rely on its own output in generating the data, similar to the scheduled sampling, it is avoiding on reliance to the ground-truth data and hence avoids the exposure bias problem.
However, adversarial generative models, in general, suffer from the reward sparsity~\cite{lin2017adversarial,che2017maximum} and mode collapse~\cite{zhang2017adversarial} problems.
Although, there are ways to avoid these two problems~\cite{shi2018towards}, studying these solutions is outside the scope of this work.

The second problem with seq2seq models is that, while the model training is done using the $\mathcal{L}_{CE}$, the model is typically evaluated during the test time using discrete and non-differentiable measures such as $\textrm{BLEU}$ and $\textrm{ROUGE}$.
This will create a mismatch between the training objective and the test objective and therefore could yield inconsistent results.
Thus, a solution that could use these measures during training of the model will inherently solve this mismatch problem.
Recently, it has been shown that both the \textit{exposure bias} and non-differentiability of evaluation measures can be addressed by incorporating techniques from reinforcement learning~\cite{bahdanau2016actor,rennie2016self,paulus2017deep,williams1992simple}.


\subsection{Reinforcement Learning}
\label{section:introrl}
In RL, a sequential Markov Decision Process (MDP) is considered, in which an agent interacts with an environment $\upvarepsilon$ over discrete time steps $t$\cite{sutton1998reinforcement}.
Let $M=(\mathcal{S}, \mathcal{A}, \mathcal{P}, R, s_0, \gamma, T)$ represent this discrete finite-horizon discounted Markov decision process, where $\mathcal{S}$ is the set of states, $\mathcal{A}$ is the set of actions, $\mathcal{P}:\mathcal{S}\times\mathcal{A}\times\mathcal{S}\rightarrow{\mathbb{R}_+}$ is the transition probability distribution, $\mathcal{R}:\mathcal{S}\times\mathcal{A}\rightarrow\mathbb{R}$ is a reward function, $s_0: \mathcal{S}\rightarrow{\mathbb{R}_+}$ is the initial state distribution, $\gamma\in[0,1]$ a discount factor, and $T$ is the horizon.

The goal of the agent is to excel at a specific task, e.g., moving an object~\cite{levine2016end,lenz2015deep}, playing an Atari game~\cite{mnih2015human}, or generating news summary~\cite{paulus2017deep,li2018actor}.
The idea is that given the environment state at time $t$ as $s_t$, the agent picks an action $\hat{y}_t\in \mathcal{A}$, according to a (typically stochastic) policy $\pi(\hat{y}_t|s_t):\mathcal{S}\times\mathcal{A}\rightarrow{\mathbb{R}_+}$ and observes a reward $r_t$ for that action\footnote{we remove the subscript $t$ whenever it is clear from the context that we are in time $t$}. The cumulative discounted sum of rewards is the
objective function optimized by policy $\pi$. 
For instance, we can consider our seq2seq conditioned RNN as a stochastic policy that generates actions (selecting the next output) and receives the task reward based on a discrete measure like $\textrm{ROUGE}$ as the return. 
The agent's goal is to maximize the expected discounted reward, $R_t = \mathbb{E}_{\pi}[\sum_{\tau=0}^{T}\gamma^{\tau}r_\tau]$, where the discounting factor $\gamma$ controls the trades off between the importance of immediate and future rewards.
Under the policy $\pi$, we can define the values of the state-action pair $Q(s_t, y_t)$ and the state $V(s_t)$ as follows:
\begin{equation}
\begin{array}{l}
Q_{\pi}(s_t,y_t)=\mathop{\mathbb{E}} [r_{t}|s=s_{t},y=y_t]\\
V_{\pi}(s_t) = \mathop{\mathbb{E}_{y \sim \pi(s)}}[Q_{\pi}(s_t,y=y_t)]
\end{array}
\label{eq:qv}
\end{equation}
\noindent
Note that the value function $V_{\pi}$ is defined over
only the states whereas $Q_{\pi}$ is defined over (state,
action) pairs. The $Q_{\pi}$ formulation is advantageous
in model-free contexts since it can be applied to {\it current} states without having access to a model of the environment.
In contrast, the $V_{\pi}$ formulation must, by necessity, be applied to {\it future} states and thus requires a model of the environment (i.e., which states and actions lead to which other future
states).
The preceding state-action function ($Q$ function for short) can be computed recursively with dynamic programming:
\begin{equation}
Q_{\pi}(s_t,y_t)=\mathop{\mathbb{E}_{s_{t^{'}}}} [r_{t}+\gamma \underbrace{\mathop{\mathbb{E}_{y_{t^{'}}\sim \pi(s_{t^{'}})}}[Q_{\pi}(s_{t^{'}},y_{t^{'}})]}_{V_{\pi}(s_{t^{'}})}]
\label{eq:viteration}
\end{equation}

Given the above definitions, we can define a function called advantage, denoted by $A_{\pi}$ relating the value function $V$ and $Q$ function as follows:
\begin{equation}
\begin{array}{ll}
A_{\pi}(s_{t},y_t) &= Q_{\pi}(s_{t},y_t)-V_{\pi}(s_{t}) \\
&=r_{t} + \gamma \mathop{\mathbb{E}_{s_{t^{'}}\sim \pi(s_{t^{'}}|s_t)}}[V_{\pi}(s_{t^{'}})] - V_{\pi}(s_t)
\end{array}
\label{eq:adv}
\end{equation}
where $\mathop{\mathbb{E}_{y\sim \pi(s)}} [A_{\pi}(s,y)]=0$ and for a deterministic policy, $y^{*}=\argmax_{y}Q(s,y)$, it follows that $Q(s,y^{*})=V(s)$, hence $A(s,y^{*})=0$. Intuitively, the value function $V$ measures how good the model could be when it is in a specific state $s$. The $Q$ function, however, measures the value of choosing a specific action when we are in such state. Given these two functions, we can obtain the advantage function which captures the importance of each action by subtracting the value of the state $V$ from the $Q$ function.
In practice, seq2seq model is used as the policy which generates actions. The definition of an action, however, will be task-specific; e.g., for a text summarization task, the action denotes choosing the next token for the summary, whereas for a question answering task, the action might be defined as the start and end index of the answer in the document. Also, the definition of the reward function could vary from one application to another. For instance, in text summarization, measures like $\textrm{ROUGE}$ and $\textrm{BLEU}$ are commonly used while in image captioning, $\textrm{CIDEr}$ and $\textrm{METEOR}$ are common. Finally, the state of the model is usually defined as the decoder output state at each time step. Therefore, the decoder output state at each time is used as the current state of the model and is used to calculate our $Q$, $V$, and advantage functions. Table~\ref{table:notation} summarizes the notations used in this paper.

\begin{table*}[t]
\centering
\caption{Notations used in this paper.}
\label{table:notation}
\begin{tabular}{|c|l|}
\hline
\multicolumn{2}{|c|}{\textbf{Seq2seq Model Parameters}}                                                                                                                                                                                                                          \\ \hline
$X$                                                                     & The sequence of input of length $T_e$, $X=\{x_1,x_2,\cdots,x_{T_e}\}.$                                                                                                                                  \\ \hline
$Y$                                                                     & The sequence of ground-truth output of length $T$, $Y=\{y_1,y_2,\cdots,y_T\}.$                                                                                                                          \\ \hline
$\hat{Y}$                                                               & The sequence of output generated by model of length $T$, $\hat{Y}=\{\hat{y}_1,\hat{y}_2,\cdots,\hat{y}_T\}.$                                                                                            \\ \hline
$T_e$                                                                   & Length of the input sequence and number of encoders.                                                                                                                                                    \\ \hline
$T$                                                                     & Length of the output sequence and number of decoders.                                                                                                                                                   \\ \hline
$d$                                                                     & Size of the input and output sequence representation.                                                                                                                                                   \\ \hline
$\mathcal{A}$                                                           & Input and output shared vocabulary.                                                                                                                                                                     \\ \hline
$h_t$                                                                   & Encoder hidden state at time $t$.                                                                                                                                                                       \\ \hline
$s_t$                                                                   & Decoder hidden state at time $t$.                                                                                                                                                                       \\ \hline
$\pi_\theta$                                                            & The seq2seq model with parameter $\theta$.                                                                                                                                                              \\ \hline
\multicolumn{2}{|c|}{\textbf{Reinforcement Learning Parameters}}                                                                                                                                                                                                                 \\ \hline
$r_t=r(s_t,y_t)$                                                        & The reward that the agent receives by taking action $y_t$ when the state of the environment is $s_t$                                                                                                   \\ \hline
$\hat{Y}$                                                               & \begin{tabular}[c]{@{}l@{}}Sets of actions that the agent is taking for a period of time $T$, $\hat{Y}=\{\hat{y}_1,\hat{y}_2,\cdots,\hat{y}_T\}$\\ This is similar to the output that the seq2seq model is generating.\end{tabular} \\ \hline
\begin{tabular}[c]{@{}c@{}}$\pi$\\ $\pi_\theta$\end{tabular}            & \begin{tabular}[c]{@{}l@{}}The policy that the agent uses to take the action.\\ Seq2seq models use  RNNs  with parameter $\theta$ for the policy.\end{tabular}                              \\ \hline
$\gamma$                                                                & Discount factor to reduce the effect of rewards from future actions.                                                                                                                                  \\ \hline
\begin{tabular}[c]{@{}c@{}}$Q(s_t,y_t)$\\ $Q_\pi(s_t,y_t)$\end{tabular} & The $Q$-value (under policy $\pi$) that shows the estimated reward of taking action $y_t$ when at state $s_t$.                                                                                    \\ \hline
$Q_{\Psi}(s_t,y_t)$                                                     & A function approximator with parameter $\Psi$ that estimates the $Q$-value given the state-action pair at time $t$.                                                                                      \\ \hline
\begin{tabular}[c]{@{}c@{}}$V(s_t)$\\ $V_\pi(s_t)$\end{tabular}         & Value function which calculates the expectation of $Q$-value (under policy $\pi$) over all possible actions.                                                                                              \\ \hline
$V_{\Psi}(s_t)$                                                         & A function approximator with parameter $\Psi$ that estimates the value function given the state at time $t$.                                                                                           \\ \hline
$A_\pi(s_t,y_t)$                                                        & \begin{tabular}[c]{@{}l@{}}Advantage function (under policy $\pi$) which defines how good a state-action pair is \\ w.r.t. the expected reward that can be received at this state.\end{tabular}                \\ \hline
$A_\Psi(s_t,y_t)$                                                       & A function approximator with parameter $\Psi$ that estimates the advantage function for the state-action pair at time $t$.                                                                                 \\ \hline
\end{tabular}
\end{table*}

\subsection{Paper Organization}
\label{section:organizaition}
In general, we define the following problem statement that we are trying to solve by combining these two different models of learning.

\textbf{Problem Statement}: \textit{Given a series of input data and a series of ground-truth outputs, train a model that:
\begin{itemize}
\item Only relies on its own output, rather than the ground-truth, to generate the results (avoiding exposure bias).
\item Directly optimize the model using the evaluation measure (avoiding mismatch between training and test measures).
\end{itemize}
}

Although recently there had been a couple of survey articles on the topic of deep reinforcement learning~\cite{arulkumaran2017brief,li2017deep}, these works heavily focused on the reinforcement learning methods and their applications in robotics and vision, while giving less emphasis to how these models could be used in a variety of other tasks. In this paper, we will summarize some of the most recent frameworks that attempted to find a solution for the above problem statement in a broad range of applications and explain how RL and seq2seq learning could benefit from each other in solving complex tasks. To this end, we will provide insights on some of the challenges and issues with the current models and how one can improve them with better RL models. The goal of this paper is to provide information about how we can broaden the power of seq2seq models with RL methods and understand challenges that exist in applying these methods to deep learning contexts. In addition, currently, there does not exist a good open-source framework for implementing these ideas. Along with this paper, we provide a library that combines state-of-the-art methods for the complex task of abstractive text summarization with recent techniques used in deep RL. The library provides a variety of different options and hyperparameters for training the seq2seq model using different RL models.
Moreover, We provide experimental results on some of the most common techniques that are explained in this paper and we encourage researchers to experiment with other hyperparameters and explore how they can use this framework to gain better performance on different seq2seq tasks.
The contributions of this paper are summarized as follows:
\begin{itemize}[leftmargin=*]
\item We provide a comprehensive summary of RL methods that are used in deep learning and specifically in the context of  training seq2seq models.
\item We summarize the challenges, advantages, and disadvantages of using different RL methods for seq2seq training.
\item We provide guidelines on how one could improve a specific RL method to obtain a better and smoother training for seq2seq models.
\item We provide an open-source library for implementing a complex seq2seq model using different RL techniques~\footnote{\url{www.github.com/yaserkl/RLSeq2Seq/}} along with experiments that aim for identifying an accurate estimate on the amount of improvement that RL algorithm provide for current seq2seq models.
\end{itemize}

This paper is organized as follows: Section~\ref{section:rl} provides details of some of the common RL techniques used in training seq2seq models.
We provide a brief introduction to different seq2seq models in Section~\ref{rlseq2seq} and later explain various RL models that can be used along with the seq2seq model training process.
We provide a summary of recent real-world applications that combine RL training with seq2seq training and in Section~\ref{rlseq} we present the framework that we implemented and discuss the details about how this framework can applied to different seq2seq problems and provide experimental results for some of the well-known RL algorithm.
Finally, in Section~\ref{section:conclusion}, we discuss conclusions of our work.
\section{Seq2seq Models and their Applications}
\label{seq2seq}
Sequence to Sequence (seq2seq) models have been an integral part of many real-world problems. From Google Machine Translation~\cite{wu2016google} to Apple's Siri speech to text~\cite{siri2017}, seq2seq models provide a clear framework to process information that is in the form of sequences. In a seq2seq model, the input and output are in the form of sequences of single units like sequence of words, images, or speech units. Table~\ref{table:seq2seq} provides a brief summary of various seq2seq models and their corresponding inputs and outputs. We also cite some of the important research papers for each application domain.

\begin{table*}
\caption{A summary of different applications of seq2seq models. In seq2seq models, the input and output are sequences of unit data. The input column provides information about the sequences of data that are fed into the model and the output column provides information about the sequences of data that the model generates as its output.}
\label{table:seq2seq}
\begin{tabular}{|l|l|l|l|l|}
\hline
\textbf{Application}                                                                          & \textbf{Problem Description}                                                                                                                           & \textbf{Input}                                                                                                                         & \textbf{Output}                                                                                                                                                                                               & \textbf{References}                                                                                                                                                                                  \\ \hline
\textbf{Machine Translation}                                                              & \begin{tabular}[c]{@{}l@{}}Translating a sentence from a\\ source language to a target\\ language\end{tabular}                                         & \begin{tabular}[c]{@{}l@{}}A sentence (sequence of words)\\ in language X (e.g., English)\end{tabular}                                  & \begin{tabular}[c]{@{}l@{}}Another sentence (sequence of\\ words) in language Y\\ (e.g., French)\end{tabular}                                                                                                  & \begin{tabular}[c]{@{}l@{}}\cite{bahdanau2014neural,sutskever2014sequence,cho2014learning}\\ \cite{luong2015effective,klein2017opennmt,chen2018best}\end{tabular}                                                                                      \\ \hline
\textbf{\begin{tabular}[c]{@{}l@{}}Text Summarization\\ Headline Generation\end{tabular}} & \begin{tabular}[c]{@{}l@{}}Summarizing a document\\ into a more concise and\\ shorter text\end{tabular}                                             & \begin{tabular}[c]{@{}l@{}}A long document like a news\\ article (sequence of words)\end{tabular}                                      & \begin{tabular}[c]{@{}l@{}}A short summary/headline\\ (sequence of words)\end{tabular}                                                                                                                        & \begin{tabular}[c]{@{}l@{}}\cite{rush2015neural,chopra2016abstractive,nallapati2017summarunner,ling2017coarse}\\ \cite{see2017get,zhou2017selective,tan2017abstractive,chen2018fast,lin2018global,hsu2018unified,xia2017deliberation}\end{tabular} \\ \hline
\textbf{Question Generation}                                                              & \begin{tabular}[c]{@{}l@{}}Generating interesting\\ questions from a text\\ document or an image\end{tabular}                                          & \begin{tabular}[c]{@{}l@{}}A piece of text (sequence of\\ words) or image (sequence\\ of layers)\end{tabular}                          & \begin{tabular}[c]{@{}l@{}}A set of questions (sequence\\ of words) related to the text\\ or image\end{tabular}                                                                                               & \cite{serban2016generating,mostafazadeh2016generating,yang2017semi,yuan2017machine}                                                                                                 \\ \hline
\textbf{Question Answering}                                                               & \begin{tabular}[c]{@{}l@{}}Given a text document or an\\ image and a question, find\\ the answer to the question\end{tabular}                          & \begin{tabular}[c]{@{}l@{}}A textual question (sequence of\\ words) or an image (sequence of\\ layers)\end{tabular}                     & \begin{tabular}[c]{@{}l@{}}A single word answer from\\ a document or the start and\\ end index of the answer in\\ the document\end{tabular}                                                                   & \cite{antol2015vqa,xiong2016dynamic,yang2016stacked}                                                                                                               \\ \hline
\textbf{Dialogue Generation}                                                              & \begin{tabular}[c]{@{}l@{}}Generate a dialogue between\\ two agents e.g., between a\\ robot and human\end{tabular}                                      & \begin{tabular}[c]{@{}l@{}}A dialogue from the first agent\\ (sequence of words) or audibles\\ (sequence of speech units)\end{tabular} & \begin{tabular}[c]{@{}l@{}}A dialogue from the second\\ agent (sequence of words) or\\ audibles (sequence of speech\\ units)\end{tabular}                                                                     & \cite{vinyals2015neural,li2015diversity,serban2016building,bordes2016learning}                                                                                                                         \\ \hline
\textbf{Semantic Parsing}                                                              & \begin{tabular}[c]{@{}l@{}}Generating automatic SQL\\queries from a given\\ human-written description\end{tabular}                                      & \begin{tabular}[c]{@{}l@{}}A human-written description\\of the query (sequence of words)\end{tabular} & \begin{tabular}[c]{@{}l@{}}The SQL command\\equivalent to that description\end{tabular}                                                                     & \cite{zhong2018seqsql,xu2017sqlnet}\\ \hline
\textbf{Image Captioning}                                                                 & \begin{tabular}[c]{@{}l@{}}Given an image, generate\\ a caption that explains the\\ content of the image\end{tabular}                                  & An image (sequence of layers)                                                                                                          & \begin{tabular}[c]{@{}l@{}}The caption (sequence of\\ words) describing that image\end{tabular}                                                                                                               & \begin{tabular}[c]{@{}l@{}}\cite{kiros2014multimodal,kiros2014unifying,chen2015mind,mao2014deep}\\ \cite{xu2015show,vinyals2015show,fang2015captions}\end{tabular} \\ \hline
\textbf{Video Captioning}                                                                 & \begin{tabular}[c]{@{}l@{}}Given a video clip, generate\\ a caption that explains the\\ content of the video\end{tabular}                              & A video (sequence of images)                                                                                                           & \begin{tabular}[c]{@{}l@{}}The caption (sequence of\\ words) describing the video\end{tabular}                                                                                                                & \cite{donahue2015long,venugopalan2014translating,venugopalan2015sequence,venugopalan2016improving}                                                                                  \\ \hline
\textbf{Computer Vision}                                                                  & \begin{tabular}[c]{@{}l@{}}Finding interesting events\\ in a video clip, e.g., predicting\\ the next action of a specific\\ object in the video\end{tabular} & A video (sequence of images)                                                                                                           & \begin{tabular}[c]{@{}l@{}}Differs from application\\ to application. For instance,\\ one might be interested in\\ determining the next action\\ of a specific object or entity\\ in the video\end{tabular} & \cite{sermanet2013overfeat,erhan2014scalable,girshick2015fast,ren2015faster}                                                                                                        \\ \hline
\textbf{Speech Recognition} & \begin{tabular}[c]{@{}l@{}}Given a segment of audible\\ input (e.g., speech),\\ convert it to text and\\ vice versa\end{tabular}                    & \begin{tabular}[c]{@{}l@{}}A speech (sequence of speech\\ units)\end{tabular}                                                          & \begin{tabular}[c]{@{}l@{}}The text of the input speech\\ (sequence of words)\end{tabular}                                                                                                                         & \cite{graves2014towards,miao2015eesen,amodei2016deep,bahdanau2016end}                                                                                                               \\ \hline
\textbf{Speech Synthesis}                                                                 & \begin{tabular}[c]{@{}l@{}}Given a segment of text it\\generates its audible sounds\end{tabular}                              & A text (sequence of words)                                                          & \begin{tabular}[c]{@{}l@{}}A speech representing \\(sequence of speech units) \\representing its audible\\sounds\end{tabular}                                                                                                                         &  \cite{ze2013statistical,fan2014tts}                                                                                  \\ \hline
\end{tabular}
\end{table*}

In recent years, different models and frameworks were proposed by researchers to achieve better and robust results on these tasks. For instance, attention-based models have been successfully applied to problems such as machine translation~\cite{luong2015effective}, text summarization~\cite{rush2015neural,chopra2016abstractive}, question answering~\cite{xiong2016dynamic}, image captioning~\cite{xu2015show}, speech recognition~\cite{bahdanau2016end}, and object detection~\cite{ba2014multiple}. In an attention-based model, at each decoding step, the previous decoder output is combined with the information from the encoder's output at a specific position to select the best decoder output. 

Although attention-based models can significantly improve the performance of seq2seq models in various tasks, in applications with large output space, it is challenging for the model to reach a desirable outcome.

On the other hand, there are more advanced models in seq2seq training like pointer-generator model~\cite{vinyals2015pointer,see2017get} and the transformers model which uses self-attention layers~\cite{vaswani2017attention}, but discussing these models is outside the scope of this paper.

Aside from these well-defined seq2seq problems, there are other related problems that partially work on the sequence of inputs but the output is not in the form of a sequence. Here are a few prominent applications that fall into this category.
\begin{itemize}[leftmargin=*]
    \item \textbf{Sentiment Analysis}~\cite{radford2017learning,dos2014deep,socher2013recursive}: The input is a sequence of words and the output is a single sentiment (positive, negative, or neutral).
    \item \textbf{Natural Language Inference}~\cite{conneau2017supervised,kim2018semantic,pan2018discourse}: Given two sentences, one as a premise and the other as a hypothesis, the goal is to classify the relationship between these two sentences into one of the entailment, neutrality, and contradiction classes.
    \item \textbf{Sentiment Role Labeling}~\cite{fitzgerald2015semantic,he2015question,marcheggiani2017simple,marcheggiani2017encoding}: Given a sentence and a predicate, the goal is to answer questions like ``who did what to whom and when and where''.
    \item \textbf{Relation Extraction}~\cite{mintz2009distant,huang2016attention,qin2018robust}: Given a sentence, the goal is to identify whether a specific relationship exists in that sentence or not. For instance, based on the sentence ``Barack Obama is married to Michelle Obama'', we can extract the ``spouse'' relationship.
    \item \textbf{Pronoun Resolution}~\cite{chen2016chinese,yin2017chinese,trinh2018simple,yin2018deep}: Given a sentence and a question about a pronoun in the sentence, the goal is to identify who that pronoun is referring to. For instance, in the sentence ``Susan cleaned Alice's bedroom for the help \textbf{she} had given'', the goal is to find who the word ``she'' is referring to.
\end{itemize}
Note that, although in these applications, only the input data is represented in terms of sequences, we still consider them to be seq2seq problems.

\subsection{Evaluation Measures}
Seq2seq models are usually trained with cross-entropy loss, i.e., Eq. (\ref{eq:cel}). However, the performance of these models is evaluated using discrete measures. 
There are various discrete measures that are used for evaluating these models and each application requires its own evaluation measure.
We briefly provide a summary of these measures according to their application context:
\begin{itemize}[leftmargin=*]
\item $\textrm{\textbf{ROUGE}}$~\footnote{\url{https://github.com/andersjo/pyrouge/}}~\cite{lin2004rouge}, $\textrm{\textbf{BLEU}}$~\footnote{\url{https://www.nltk.org/_modules/nltk/translate/bleu_score.html}}~\cite{papineni2002bleu}, $\textrm{\textbf{METEOR}}$~\footnote{\url{http://www.cs.cmu.edu/~alavie/METEOR/}}~\cite{banerjee2005meteor}: These are three of the most commonly used measures in applications such as machine translation, headline generation, text summarization, question answering, dialog generation, and other applications that require evaluation of text data.
$ROUGE$ measure finds the common unigram ($ROUGE$-1), bigram ($ROUGE$-2), and largest common substring (LCS) ($ROUGE$-L) between the ground-truth text and the output generated by the model and calculate respective precision, recall, and F-score for each measure.
$BLEU$ works similar to $ROUGE$ but through a modified precision calculation, it inclines to provide higher scores to outputs that are closer to human judgement.
In a similar manner, $METEOR$ uses the harmonic mean of unigram precision and recall and it gives higher importance to recall than the precision.
Although these methods are designed to work for all text-based applications, $METEOR$ is more often used in machine translation tasks, while $ROUGE$ and $BLEU$ are mostly used in text summarization, question answering, and dialog generation.

\item $\textrm{\textbf{CIDEr}}$~\footnote{\url{https://github.com/vrama91/cider}}~\cite{vedantam2015cider}, $\textrm{\textbf{SPICE}}$~\footnote{\url{http://www.panderson.me/spice/}}~\cite{spice2016}: $CIDEr$ is frequently used in image and video captioning tasks in which having captions that have higher human judgement scores is more important. Using sentence similarity, the notions of grammaticality, saliency, importance, accuracy, precision, and recall are inherently captured by these metrics.

$SPICE$ is a recent evaluation metric proposed for image captioning that tries to solve some of the problems of $CIDEr$ and $METEOR$ by mapping the dependency parse trees of the caption to the semantic scene graph (contains objects, attributes of objects, and relations) extracted from the image. Finally, it uses the F-score that is calculated using the tuples of the generated and ground-truth scene graphs to provide the caption quality score.

\item \textbf{Word Error Rate} ($\textrm{\textbf{WER}}$): This measure, which is mostly used in speech recognition, finds the number of substitutions, deletions, insertions, and corrections required to change the generated output to the ground-truth and combines them to calculate the $WER$.

\end{itemize}

\subsection{Datasets}
In this section, we briefly describe some of the  datasets that are commonly used in various seq2seq models.
We provide a short list of some of the most common datasets that are used in various seq2seq applications as follows:
\begin{itemize}[leftmargin=*]
\item \textbf{Machine Translation}: The most common dataset used for Machine Translation task is the \textbf{WMT'14}~\footnote{\url{http://www.statmt.org/wmt14/translation-task.html}} dataset which contains 850M words from English-French parallel corpora of UN (421M words), Europarl (61M words), news commentary (5.5M words), and two crawled corpora of 90M and 272.5M words. The data pre-processing for this dataset is usually done following the code~\footnote{\url{http://www-lium.univ-lemans.fr/~schwenk/cslm_joint_paper/}} provided by Axelrod \textit{et al.}~\cite{axelrod2011domain} .

\item \textbf{Text Summarization}: One of the main datasets used in text summarization is the \textbf{CNN-Daily Mail} dataset~\cite{hermann2015teaching} which is part of the DeepMind Q\&A Dataset~\footnote{\url{https://cs.nyu.edu/~kcho/DMQA/}} and contains around 287K news articles along with 2 to 4 highlights (summary) for each news article~\footnote{For downloading and pre-processing please refer to: \url{https://github.com/abisee/cnn-dailymail}}.
Recently, another dataset, called \textbf{Newsroom}, was released by Connected Experiences Lab~\footnote{\url{https://summari.es/}}~\cite{Grusky2018Newsroom} which contains 1.3M news articles and various metadata information such as the title and summary of the news.
The document summarization challenge~\footnote{\url{https://duc.nist.gov/data.html}} also provides some datasets for text summarization. More specifically, in this dataset, \textbf{DUC-2003} and \textbf{DUC-2004} which contain 500 news articles (each paired with 4 different human-generated reference summaries) from the New York Times and Associated Press Wire services, respectively. Due to the small size of this dataset, researchers usually use this dataset only for evaluation purposes.

\item \textbf{Headline Generation}: Headline generation is similar to the task of text summarization and typically all the datasets that are used in text summarization will be useful in headline generation, too.
There is a big dataset which is called \textbf{Gigaword}~\cite{graff2003english} and contains more than 8M news articles from multiple news agencies like New York Times and Associate Press. However, this dataset is not freely available and researchers are required to buy the license to be able to use it though one can still find pre-trained models on different tasks using this dataset~\footnote{\url{http://opennmt.net/Models/}}.

\item \textbf{Question Answering and Question Generation}: The \textbf{CNN-Daily Mail} dataset was originally designed for question answering and is one of the earliest datasets that is available for tackling this problem.
However, recently two large-scale datasets that are solely designed for this problem were released.
Stanford Question Answering Dataset (\textbf{SQuAD})~\footnote{\url{https://rajpurkar.github.io/SQuAD-explorer/}}(1.0 and 2.0)~\cite{rajpurkar2016squad,rajpurkar2018know} is a dataset for reading comprehension and contains more than 100K pairs of questions and answers collected by crowd-sourcing over a set of Wikipedia articles. The answer to each question is a segment which identifies the start and end indices of the answer within the article. The second dataset is called \textbf{TriviaQA}~\footnote{\url{http://nlp.cs.washington.edu/triviaqa/}}~\cite{joshi2017triviaqa}, and similar to \textbf{SQuAD}, it is designed for reading comprehension and question answering task. This dataset contains 650K triples of questions, answers, and evidences (which help to find the answer).

\item \textbf{Dialogue Generation}: The dataset for this problem usually comprises of dialogues between different people. The \textbf{OpenSubtitles} dataset~\footnote{\url{http://opus.nlpl.eu/OpenSubtitles.php}}~\cite{tiedemannnews}, \textbf{Movie Dialog dataset}~\footnote{\url{http://fb.ai/babi}}~\cite{dodge2015evaluating}, and \textbf{Cornell Movie Dialogues} Corpus~\footnote{\url{http://www.cs.cornell.edu/~cristian/Cornell_Movie-Dialogs_Corpus.html}}~\cite{danescu2011chameleons} are three examples of these types of datasets.
\textbf{OpenSubtitles} contains conversations between movie characters for more than 20K movies in 20 languages. The \textbf{Cornell Movie Dialogues} corpus contains more than 220K dialogues between more than 10K movie characters.

\item \textbf{Semantic Parsing}: Recently Zhong \textit{et al.}~\cite{zhong2018seqsql} released a dataset called \textbf{WikiSQL}~\footnote{\url{https://github.com/salesforce/WikiSQL}} for this problem which contains 80654 hand-annotated questions and SQL queries distributed across 24241 tables from Wikipedia. Although this is not the only dataset for this problem but it offers a larger set of examples from other datasets such as \textbf{WikiTableQuestion}~\footnote{\url{https://nlp.stanford.edu/software/sempre/wikitable/}}~\cite{pasupat2015compositional} and \textbf{Overnight}~\cite{wang2015building}.

\item \textbf{Sentiment Analysis}: For this application, \textbf{Amazon product review}~\footnote{\url{http://jmcauley.ucsd.edu/data/amazon/}}~\cite{mcauley2015inferring} dataset is one of the largest dataset which contains more than 82 million product reviews from May 1996 to July 2014 in its de-duplicated version. Another big dataset for this task is the \textbf{Stanford Sentiment Treebank (SSTb)}~\footnote{\url{https://nlp.stanford.edu/sentiment/}}~\cite{socher2013recursive}, which includes fine grained sentiment labels for 215,154 phrases in the parse trees of 11,855 sentences.

\item \textbf{Natural Language Inference}: \textbf{Stanford Natural Language Inference} (SNLI)~\footnote{\url{https://nlp.stanford.edu/projects/snli/}}~\cite{snli:emnlp2015} is the standard dataset for this task which contains 570K human-written English sentence pairs manually labeled for the three classes entailment, contradiction, and neutral. The \textbf{Multi-Genre Natural Language Inference} (MultiNLI)~\footnote{\url{https://www.nyu.edu/projects/bowman/multinli/}}~\cite{williams2018broad} corpus is another new dataset which is collected through crowd-sourcing and contains 433K sentence pairs annotated with textual entailment information.

\item \textbf{Semantic Role Labeling}: \textbf{Proposition Bank} (PropBank)~\footnote{\url{http://propbank.github.io/}}~\cite{palmer2005proposition} is the standard dataset for this task which contains a corpus of text annotation with information about basic semantic propositions in seven different languages.

\item \textbf{Relation Extraction}: \textbf{Freebase}~\footnote{\url{https://old.datahub.io/dataset/freebase}}~\cite{finkel2005incorporating} is a huge dataset containing billions of triples: the entity pair and the specific relationship between them which are selected from the New York Times corpus (NYT).

\item \textbf{Pronoun Resolution}: The \textbf{OntoNotes 5.0} dataset~\footnote{\url{https://catalog.ldc.upenn.edu/LDC2013T19}} is the standard dataset for this task.
Specifically, researchers use the Chinese portion of this dataset to do the pronoun resolution in Chinese~\cite{chen2016chinese,yin2017chinese,yin2018deep}.

\item \textbf{Image Captioning}: There are two datasets that are mainly used in image captioning. The first one is the \textbf{COCO} dataset~\footnote{\url{http://cocodataset.org/}}~\cite{lin2014microsoft} which is designed for object detection, segmentation, and image captioning. This dataset contains around 330K images amongst which 82K images are used for training and 40K used for validation in image captioning. Each image has five ground-truth captions. \textbf{SBU}~\cite{ordonez2011im2text} is another dataset which consists of 1M images from Flickr and contains descriptions provided by image owners when they uploaded the images to Flickr.

\item \textbf{Video Captioning}: For this problem, \textbf{MSR-VTT}~\footnote{\url{http://ms-multimedia-challenge.com/2017/challenge}}~\cite{xu2016msr} and \textbf{YouTube2Text/MSVD}~\footnote{\url{http://www.cs.utexas.edu/users/ml/clamp/videoDescription/}}~\cite{chen2011collecting} are two of the widely used datasets. MSR-VTT consists 10K videos from a commercial video search engine each containing 20 human annotated captions and YouTube2Text/MSVD which has 1970 videos each containing on an average 40 human annotated captions.

\item \textbf{Image Classification}: The most popular dataset in computer vision is the \textbf{MNIST} dataset~\footnote{\url{http://yann.lecun.com/exdb/mnist/}}~\cite{lecun1998gradient}.
This dataset consists of handwritten digits and contains a training set of 60K examples and a test set of 10K examples.
Aside from this dataset, there is a huge list of datasets that are used for various computer vision problems and explaining each of them is beyond the scope of this paper~\footnote{Please refer to this link for a comprehensive list of datasets that are used in computer vision: \url{http://riemenschneider.hayko.at/vision/dataset/}}.

\item \textbf{Speech Recognition}: \textbf{LibriSpeech ASR Corpus}~\footnote{\url{http://www.openslr.org/12/}}~\cite{panayotov2015librispeech} is one of the main datasets used for the speech recognition task. This dataset is free and is composed of 1000 hours of segmented and aligned 16kHz English speech which is derived from audiobooks. \textbf{Wall Street Journal} (WSJ) also has two Continuous Speech Recognition corpora containing 70 hours of speech and text from a corpus of Wall Street Journal news text. However, unlike the LibriSpeech dataset, this dataset is not freely available and researchers have to buy a license to use it. Similar to the WSJ dataset, \textbf{TIMIT}~\footnote{\url{https://catalog.ldc.upenn.edu/ldc93s1}} is another dataset containing the read speech data. It contains time-aligned orthographic, phonetic, and word transcriptions of recordings for 630 speakers of eight major dialects of American English in which each of them are reading ten phonetically sentences.

\end{itemize}
\section{ Reinforcement Learning Methods}
\label{section:rl}
In reinforcement learning, the goal of an agent interacting with an environment is to maximize the expectation of the reward that it receives from the actions. 
Therefore, the focus is on maximizing one of the following objectives:
\begin{equation}
\maximize\ \mathop{\mathbb{E}_{\hat{y}_1,\cdots,\hat{y}_T \sim {\pi_{\theta}(\hat{y}_1,\cdots,\hat{y}_T)}}}[r(\hat{y}_1,\cdots,\hat{y}_T)]
\label{eq:pg}
\end{equation}
\begin{equation}
\maximize_{y}\ A_\pi(s_t, y_t)
\label{eq:aargmax}
\end{equation}
\begin{equation}
\maximize_{y}\ A_\pi(s_t, y_t) \rightarrow Maximize_{y}\ Q_\pi(s_t, y_t)
\label{eq:qargmax}
\end{equation}
There are various ways in which one can solve this problem.
In this section, we explain the solutions in detail and provide their strengths and weaknesses.
Different methods aim solving this problem by trying one of the following approaches:
(i) solve this problem through Eq. (\ref{eq:pg}); (ii) solve the expected discounted reward $\mathop{\mathbb{E}}[R_t = \sum_{\tau=t}^{T}\gamma^{\tau-t}r_{\tau}]$; (iii) solve it by maximizing the advantage function (Eq. (\ref{eq:aargmax})); and (iv) solve it by maximizing $Q$ function using Eq. (\ref{eq:qargmax}). Most of these methods are suitable choices for improving the performance of seq2seq models, but depending on the approach that is chosen for training the reinforced model, the training procedure for seq2seq model also changes. The first and one of the simplest algorithms that will be discussed in this section is the Policy Gradient (PG) method which aims to solve Eq. (\ref{eq:pg}). Section~\ref{section:ac} discusses Actor-Critic (AC) methods which improve the performance of PG models by solving Eq. (\ref{eq:aargmax}) through Eq. (\ref{eq:viteration}) expansion on $Q$-function. Section~\ref{section:ql} discusses $Q$-learning models that aim at maximizing the $Q$ function (Eq. (\ref{eq:qargmax})) to improve the PG and AC models. Finally, Section~\ref{section:advq} will provide more details about some of the recent models which improve the performance of $Q$-learning models.

\subsection{Policy Gradient}
\label{section:pg}
In all reinforcement algorithms, an agent takes some action according to a specific policy $\pi$. The definition of a policy varies according to the specific application that is being considered. For instance, in text summarization, the policy is a language model $p(y|X)$ that, given input $X$, tries to generate the output $y$. Now, let us assume that our agent is represented by an RNN and takes actions from a policy $\pi_\theta$\footnote{In seq2seq model, this represents $\pi_{\theta}(y_t|\hat{y}_{t-1},s_{t},c_{t-1})$ in Eq. (\ref{eq:seq2seqp})}. In a deterministic environment, where the agent takes discrete actions, the output layer of the RNN is usually a softmax function and it generates the output from this layer. In Teacher Forcing, a set of ground-truth sequences are given and the actions are chosen according to the current policy during training and the reward is observed only at the end of the sequence or when an End-Of-Sequence (EOS) signal is seen. Once the agent reaches the end of sequence, it compares the sequence of actions from the current policy ($\hat{y}_t$) against the ground-truth action sequence ($y_t$) and calculate a reward based on any specific evaluation metric. The goal of training is to find the parameters of the agent in order to maximize the expected reward. This loss is defined as the negative expected reward of the full sequence:
\begin{equation}
\begin{array}{c}
\mathcal{L}_{\theta} = -\mathop{\mathbb{E}_{\hat{y}_1,\cdots,\hat{y}_T \sim {\pi_{\theta}(\hat{y}_1,\cdots,\hat{y}_T)}}}[r(\hat{y}_1,\cdots,\hat{y}_T)]
\end{array}
\end{equation}
where $\hat{y}_t$ is the action chosen by the model at time $t$ and $r(\hat{y}_1,\cdots,\hat{y}_T)$ is the reward associated with the actions $\hat{y}_1,\cdots,\hat{y}_T$. Usually, in practice, one will approximate this expectation with a single sample from the distribution of actions acquired by the RNN model. Hence, the derivative for the above loss function is given as follows:
\begin{equation}
\nabla_\theta \mathcal{L}_\theta = -\mathop{\mathbb{E}}_{\hat{y}_{1\cdots T}\sim \pi_\theta}[\nabla_\theta\log{\pi_\theta(\hat{y}_{1\cdots T})}r(\hat{y}_{1\cdots T})]
\end{equation}
Using the chain rule, this equation can be re-written as follows~\cite{zaremba2015reinforcement}:
\begin{equation}
\nabla_\theta \mathcal{L}_\theta = \frac{\partial \mathcal{L}_{\theta}}{\partial \theta} = \sum_{t}\frac{\partial \mathcal{L}_{\theta}}{\partial o_t}\frac{\partial o_t}{\partial \theta}
\end{equation}

where $o_t$ is the input to the softmax function. The gradient of the loss $\mathcal{L}_\theta$ with respect to $o_t$ is given by~\cite{zaremba2015reinforcement,williams1992simple}:
\begin{equation}
\frac{\partial \mathcal{L}_{\theta}}{\partial o_t} =\Big(\pi_\theta(y_{t}|\hat{y}_{t-1}, s_{t},c_{t-1}) - \textbf{1}(\hat{y}_{t})\Big)(r(\hat{y}_1,\cdots,\hat{y}_T) - r_b)
\label{eq:oloss}
\end{equation}
where $\textbf{1}(\hat{y}_{t})$ is the 1-of-$|\mathcal{A}|$ representation of the ground-truth output and $r_b$ is a baseline reward and could be any value as long as it is not dependent on the parameters of the RNN network. This equation is very similar to the gradient of a multi-class logistic regression. In logistic regression, the cross-entropy gradient is the difference between the prediction and the actual 1-of-$|\mathcal{A}|$ representation of the ground-truth output:
\begin{equation}
\begin{array}{l}
\frac{\partial \mathcal{L}_{\theta}^{CE}}{\partial o_t} = \pi_\theta(y_{t}|y_{t-1}, s_{t},c_{t-1}) - \textbf{1}(y_{t})
\end{array}
\label{eq:celr}
\end{equation}
Note that, in Eq. (\ref{eq:oloss}), the generated output from the model is used as a surrogate ground-truth for the output distribution, while, in Eq. (\ref{eq:celr}), the ground-truth is used to calculate the gradient.

The goal of the baseline reward is to force the model to select actions that yield a reward $r>r_b$ and discourage those that have reward $r<r_b$. Since only one sample is being used to calculate the gradient of the loss function, it is shown that, having this baseline would reduce the variance of the gradient estimator~\cite{williams1992simple}. If the baseline is not dependent on the parameters of the model $\theta$, Eq. (\ref{eq:oloss}) is an unbiased estimator. To prove this, we simply need to show that adding the baseline reward $r_b$ does not have any effect on the expectation of loss:
\begin{equation}
\begin{array}{ll}
&\mathop{\mathbb{E}}_{\hat{y}_{1\cdots T}\sim \pi_\theta}[\nabla_\theta\log{\pi_\theta(\hat{y}_{1\cdots T})}r_b] = 
r_b\sum_{\hat{y}_{1\cdots T}}\nabla_\theta \pi_\theta(\hat{y}_{1\cdots T}) \\ \\ 
&~~~~ = r_b \nabla_\theta \sum_{\hat{y}_{1\cdots T}}\pi_\theta(\hat{y}_{1\cdots T}) = 
r_b \nabla_\theta 1  = 0
\end{array}
\label{eq:proof}
\end{equation}

\begin{figure}
\centering
\includegraphics[width=0.99\columnwidth]{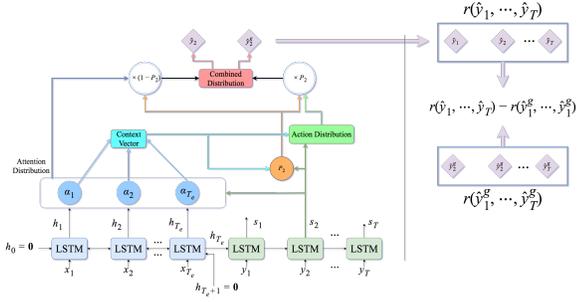}
\caption{A simple attention-based pointer-generation seq2seq model with Self-Critic reward.
At each decoding step, the context vector for that decoder is calculated and combined with the decoder output to get the action distribution.
In pointer-generation model, the attention distribution is further combined with the action distribution through switches called pointers to get the final distribution over the actions.
From each output distribution, a specific action $\hat{y}_2$ is sampled and the greedy action $\hat{y}^{g}_2$ is extracted.
The difference of the rewards from sampling and greedy sequence is used to update the loss function.
}
\label{fig:selfcritic}
\end{figure}

This algorithm is called $REINFORCE$~\cite{williams1992simple} and is a simple yet elegant policy gradient algorithm for seq2seq problems. One of the challenges with this method is that the model suffers from high variance since only one sample is used for training at each time step. To alleviate this problem, at each training step, one can sample $N$ sequences of actions and update the gradient by averaging over all these $N$ sequences as follows:
\begin{equation}
\begin{array}{l}
\mathcal{L}_\theta = \frac{1}{N} \sum_{i=1}^{N} \sum_t \log{\pi_\theta(\hat{y}_{i,t}|\hat{y}_{i,t-1}, s_{i,t},c_{i,t-1})}\times\\
\Big(r(\hat{y}_{i,1},\cdots,\hat{y}_{i,T}) - r_b\Big)
\end{array}
\label{eq:olossN}
\end{equation}
Having this, the baseline reward could be set to be the mean of the $N$ rewards that are sampled, i.e., $r_b = 1/N \sum_{i=1}^{N} r(\hat{y}_{i,1},\cdots,\hat{y}_{i,T})$.
Algorithm~\ref{alg:reinforce} shows how this method works. 

As another solution to reduce the variance of the model, \textbf{Self-Critic} (SC) models are proposed~\cite{rennie2016self}. In these SC models, rather than estimating the baseline using current samples, the output of the model obtained by a greedy-search (the output at the time of inference) is used as the baseline. Hence, the sampled output of the model is used as $\hat{y}_t$ and greedy selection of the final output distribution is used for $\hat{y}^{g}_t$ where the superscript $g$ indicates greedy selection. Following this mechanism, the new objective for the REINFORCE model would become as follows:
\begin{equation}
\begin{array}{l}
\mathcal{L}_\theta = \frac{1}{N}\sum_{i=1}^{N}  \sum_t \log{\pi_\theta(\hat{y}_{i,t}|\hat{y}_{i,t-1}, s_{i,t},c_{i,t-1})}\times\\
\Big(r(\hat{y}_{i,1},\cdots,\hat{y}_{i,T}) - r(\hat{y}^{g}_{i,1},\cdots,\hat{y}^{g}_{i,T})\Big)
\end{array}
\label{eq:scloss}
\end{equation}
Fig~\ref{fig:selfcritic} shows how an attention-based pointer-generator seq2seq model can be used to extract the reward and its baseline in Self-Critic model.

\begin{algorithm}[t]
\begin{algorithmic}
\footnotesize
\State \textbf{Input}: Input sequences ($X$), ground-truth output sequences ($Y$),
\State and (preferably) a pre-trained policy ($\pi_\theta$).
\State \textbf{Output}: Trained policy with REINFORCE.
\State \textbf{Training Steps}:
\While{not converged}
    \State Select a batch of size $N$ from $X$ and $Y$.
    \State Sample $N$ full sequence of actions:
    \State $\{\hat{y}_1,\cdots,\hat{y}_T\sim \pi_\theta(\hat{y}_1,\cdots,\hat{y}_T)\}_{1}^{N}$.
    \State Observe the sequence reward and calculate the baseline $r_b$.
    \State Calculate the loss according to Eq. (\ref{eq:olossN}).
    \State Update the parameters of network $\theta \leftarrow \theta + \alpha \nabla_{\theta} \mathcal{L}_{\theta}$.
\EndWhile
\State \textbf{Testing Steps}:
\For{batch of input and output sequences $X$ and $Y$}
	\State Use the trained model and Eq. (\ref{eq:inf}) to sample the output $\hat{Y}$.
	\State Evaluate the model using a performance metric, e.g. $ROUGE_{l}$.
\EndFor
\end{algorithmic}
\caption{REINFORCE algorithm}
\label{alg:reinforce}
\end{algorithm}

The second problem with this method is that the reward is only observed after the full sequence of actions is sampled.
This might not be a pleasing feature for most of the seq2seq models.
If we see the partial reward of a given action at time $t$, and the reward is bad, the model needs to select a better action for the future to maximize the reward. 
However, in the REINFORCE algorithm, the model is forced to wait till the end of the sequence to observe its performance.
Therefore, the model often generates poor results or takes longer to converge. This problem magnifies especially in the beginning of the training phase where the model is initialized randomly and thus selects arbitrary actions. To alleviate this problem to a certain extent, Ranzato \textit{et al.}~\cite{ranzato2015sequence} suggested to pre-train the model for a few epochs using the cross-entropy loss and then slowly switch to the REINFORCE loss. Finally, as another way to solve the high variance problem of the REINFORCE algorithm, importance sampling~\cite{jie2010connection,liu2017improved} can also be used. The basic underlying idea of using the importance sampling with REINFORCE algorithm is that rather than sampling sequences from the current model, one can sample them from an old model and use them to calculate the loss.

\subsection{Actor-Critic Model}
\label{section:ac}
As mentioned in Section~\ref{section:pg}, adding a baseline reward is a necessary component of the PG algorithm in order to reduce the variance of the model. In PG, the average reward from multiple samples in the batch was used as the baseline reward for the model. In Actor-Critic (AC) model, the goal is to train an estimator for calculating the baseline reward. For computing this quantity, AC models try to maximize the advantage function through Eq. (\ref{eq:viteration}) extension. Therefore, these methods are also called Advantage Actor-Critic (A2C) models.

In these models, the goal is to solve this problem using the following objective:
\begin{equation}
\begin{array}{l}
A_{\pi}(s_{t},y_{t}) = Q_{\pi}(s_{t},y_{t})-V_{\pi}(s_{t}) = \\
r_{t} + \gamma \mathop{\mathbb{E}_{s_{t^{\prime}}\sim \pi(s_{t^{\prime}}|s_t)}}[V_{\pi}(s_{t^{\prime}})] - V_{\pi}(s_t)
\end{array}
\end{equation}
Similar to the PG algorithm, to avoid the expensive inner expectation calculation, we can only sample once and approximate advantage function as follows:
\begin{equation}
\begin{array}{l}
A_{\pi}(s_{t},y_{t}) \approx r_t + \gamma V_{\pi}(s_{t^{\prime}}) - V_\pi(s_t)
\end{array}
\label{eq:simpleadv}
\end{equation}
Now, in order to estimate $V_{\pi}(s)$, a function approximator can be used to approximate the value function.
In AC, neural networks is typically used as the function approximator for the value function.
Therefore, we fit a neural network $V_{\pi}(s;{\Psi})$ with parameters $\Psi$ to approximate the value function.
Now, if we consider $r_t + \gamma V_{\pi}(s_{t^{\prime}})$ as the expectation of reward-to-go at time $t$, $V_\pi(s_t)$ could play as a surrogate for the baseline reward.
Similar to the PG, the variance of the model would be high since only one sample is used to train the model.
Therefore, the variance can be reduced using multiple samples.
In the AC model, the Actor (our policy, $\theta$) provides samples (policy states at time $t$ and $t+1$) for the Critic (neural network estimating value function, $V_{\pi}(s;{\Psi})$) and the Critic returns the estimation to the Actor, and finally, the Actor uses these estimations to calculate the advantage approximation and update the loss according to the following equation:
\begin{equation}
\begin{array}{l}
\mathcal{L}_\theta = \frac{1}{N}\sum_{i=1}^{N} \sum_t \log{\pi_\theta(\hat{y}_{i,}|\hat{y}_{i,t-1}, s_{i,t},c_{i,t-1})}A_{\Psi}(s_{i,t},y_{i,t})
\end{array}
\label{eq:acloss}
\end{equation}

\begin{figure}
\centering
\includegraphics[width=0.99\columnwidth]{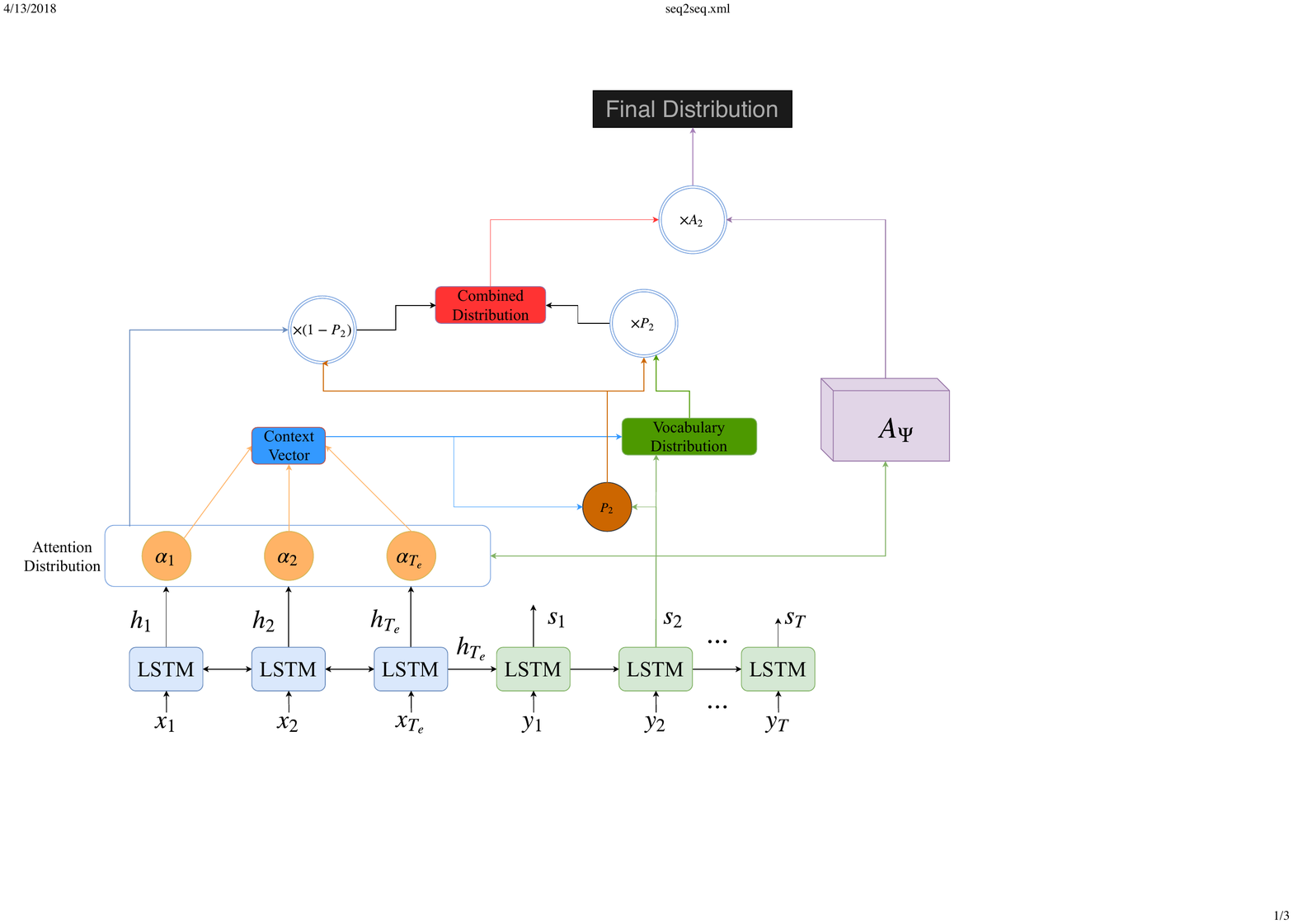}
\caption{A simple Actor-Critic model with an attention-based pointer-generation seq2seq model as the Actor. The Critic model is shown on the right side of the picture with a purple box.
The purple box $A_{\Psi}$, which represents the Critic model, takes as input the decoder output at time $t=2$, i.e., $s_2$, and estimate the advantage values through either (value function estimation, DQN, DDQN, or dueling net) for each action.
}
\label{fig:rlseq}
\end{figure}

Therefore, in the AC models, the inference at each time $t$ would be as follows:
\begin{equation}
\argmax_{y} \pi_\theta(\hat{y}_{t}|\hat{y}_{t-1}, s_{t},c_{t-1})A_{\Psi}(s_{t},y_{t})
\label{eq:acinf}
\end{equation}
Fig.~\ref{fig:rlseq} provides an illustration of how this model works at one of the decoding steps.

\subsubsection{Training Critic Model}
As mentioned in the previous section, the Critic is a function estimator which tries to estimate the expected reward-to-go for the model at time $t$. Therefore, training the Critic is basically a regression problem. Usually, in AC models, a neural network is used as the function approximator and the value function is trained using the mean square error:
\begin{equation}
\mathcal{L}(\Psi) = \frac{1}{2}\sum_i || V_{\Psi}(s_i)-v_i ||^2
\label{eq:criticloss}
\end{equation}
where $v_i=\sum_{t^{\prime}=t}^{T}r(s_{i,t^{\prime}},y_{i,t^{\prime}})$ is the true reward-to-go at time $t$.
During training the Actor model, we collect $(s_i,v_i)$ pairs and pass them to the Critic model to train the estimator.
This model is called \textit{on-policy AC}, which refers to the fact that the samples are collected at the current time to train the Critic model.
However, the samples that are passed to the Critic will be correlated to each other which causes poor generalization for the estimator.
These methods could be turn to off-policy by collecting training samples into a memory buffer and select mini-batches from this memory buffer and train the Critic network.
Off-policy AC provides a better training due to avoiding the correlation of samples that exists in the on-policy methods.
Therefore, most the models that we discuss in this paper are primarily off-policy and use a memory buffer for training the Critic model.

Algorithm~\ref{alg:ac} shows the batch Actor-Critic algorithm since for training the Critic network, we use a batch of state-rewards pair.
In the online AC algorithm, the Critic network is simply updated using just one sample and, as expected, online AC algorithm has a higher variance due to reliance on one sample for training the network.
To alleviate this problem for online AC, we can use Synchronous Advantage AC learning or Asynchronous Advantage AC (A3C) learning~\cite{mnih2016asynchronous}.
In the synchronous approach, $N$ different threads are used to train the model and each thread performs online AC for one sample and at the end of the algorithm, the gradient of these $N$ threads is used to update the gradient of the Actor model.
In the more widely used A3C algorithm, as soon as a thread calculates $\theta$, it will send the update to other threads and other threads use the updated $\theta$ to train the model.
A3C is an on-policy method with multi-step returns while there are other methods like Retrace~\cite{munos2016safe}, UNREAL~\cite{jaderberg2016reinforcement}, and Reactor~\cite{gruslys2017reactor} which provide the off-policy variations of this model by using the memory buffer.
Also, ACER~\cite{wang2016sample} mixes on-policy (from current run) and off-policy (from memory) to train the Critic network.

\begin{algorithm}[t]
\begin{algorithmic}
\footnotesize
\State \textbf{Input}: Input sequences ($X$), ground-truth output sequences ($Y$),
\State and (preferably) a pre-trained Actor model ($\pi_\theta$).
\State \textbf{Output}: Trained Actor and Critic models.
\State \textbf{Training Steps}:
\State Initialize the Actor (Seq2seq) model, $\pi_\theta$.
\State Initialize the Critic (ValueNet) model, $V_\Psi$.
\While{not converged}
    \State \textbf{Training Actor}:
    \State Select a batch of size $N$ from $X$ and $Y$.
    \State Sample $N$ full sequences of actions based on the Actor.
    \State model, $\pi_\theta$.
    \For{$n=1,\cdots,N$}
        \For{$t=1,\cdots,T$}
            \State Calculate the true (discounted) reward-to-go:
            \State $v_t=\sum_{t^{\prime}=t}^{T}\gamma^{t^{\prime}-t}r(s_{i,t^{\prime}},y_{i,t^{\prime}})$.
            \State Store training pairs for Critic: $(s_t,v_t)$.
        \EndFor
    \EndFor
    \State
    \State \textbf{Training Critic}:
    \State Select a batch of size $N_c$ from the pool of state-rewards pairs.
    \State collected from Actor.
    \For{$n=1,\cdots,N_c$}
        \State Collect the value estimates $\hat{v}_n$ from $V_\Psi$ for each
        \State state-rewards pair.
    \EndFor
     \State Minimize the Critic loss using Eq. (\ref{eq:criticloss}).
    \State
    \State \textbf{Updating Actor}:
    \State Use the estimated value for $V_{\Psi}(s_{t})$ and $V_{\Psi}(s_{t^{\prime}})$
    \State to calculate the loss using Eq. (\ref{eq:acloss}).
    \State Update parameters of the model using $\theta \leftarrow \theta + \alpha \nabla_{\theta} \mathcal{L}(\theta)$.
\EndWhile
\end{algorithmic}
\caption{Batch Actor-Critic Algorithm}
\label{alg:ac}
\end{algorithm}

In general, AC models usually have low variance due to the batch training and the use of critic as the baseline reward, but they are not unbiased if the critic is erroneous and makes a lot of mistakes. As mentioned in Section~\ref{section:pg}, PG algorithm has high variance but it provides an unbiased estimator. Now, if the PG and AC model are combined, we will likely be ending up with a model that has no bias and low variance.
This idea comes from the fact that, for deterministic policies (like seq2seq models), a partially observable loss could be driven by using the $Q$-function as follows~\cite{bahdanau2016actor,sutton2000policy}:
\begin{equation}
\begin{array}{l}
\mathcal{L}_\theta = \frac{1}{N}\sum_{i=1}^{N} \sum_t \log{\pi_\theta(\hat{y}_{i,t}|\hat{y}_{i,t-1}, s_{i,t},c_{i,t-1})}\times\\
\Big(Q_{\Psi}(s_{i,t}) - V_{\Psi^{\prime}}(s_{i,t})\Big)
\end{array}
\label{eq:pgac}
\end{equation}
However, this model requires training two different networks for $Q_{\Psi}$ function and $V_{\Psi^{\prime}}$ function as the baseline.
Note that the same model cannot be used to estimate both $Q$ function and value function since the estimator will not be an unbiased estimator anymore~\cite{tucker2018the}.
As yet another solution to create a trade-off between the bias and variance in AC, Schulman \textit{et al.}~\cite{schulman2015high} proposed the Generalized Advantage Estimation ($GAE$) model as follows:

\begin{equation}
A^{GAE}_{\Psi}(s_t,y_t) = \sum_{i=t}^{T}(\gamma \lambda)^{i-t}\Big(r(s_{i},y_{i})+\gamma V_{\Psi}(s_{i+1})-V_{\Psi}(s_{i})\Big)
\label{eq:gae}
\end{equation}

where $\lambda$ controls the trade-off between the bias and variance such that large values of $\lambda$ yield to larger variance and lower bias, while small values of $\lambda$ do the opposite.

\subsection{Actor-Critic with Q-Learning}
\label{section:ql}
As mentioned in the previous section, the value function is used to maximize the advantage function.
As an alternative to solve the maximization of advantage estimates, we can try to solve the following objective function:
\begin{equation}
Maximize_{y}\ A_\pi(s_t, y_t) \rightarrow Maximize_{y}\ Q_\pi(s_t, y_t) - \underbrace{V_\pi(s_t)}_{0}
\label{eq:qmax}
\end{equation}
This is true since we are trying to find the actions that maximize the advantage estimate and since value function does not rely on the actions, we can simply remove them from the maximization objective.
Therefore, the advantage maximization problem is simplified to $Q$ function estimation problem.
This method is called $Q$-learning and it is one of the most commonly used algorithms for RL problems.
The $Q$-learning is called to a family of off-policy algorithm used to learn a $Q$-function.
Similar to this method, SARSA algorithm~\cite{sutton2018reinforcement} is an on-policy algorithm for calculating the $Q$-function.
The major difference between SARSA and $Q$-Learning, is that the maximum reward for the next state is not necessarily used for updating the $Q$-values.
In $Q$-learning, the Critic tries to provide an estimation for the $Q$-function. Therefore, given that the policy $\pi_\theta$ is being used, our goal is to maximize the following loss at each training step:
\begin{equation}
\begin{array}{l}
\mathcal{L}_\theta = \frac{1}{N}\sum_{i=1}^{N} \sum_t \log{\pi_\theta(\hat{y}_{i,t}|\hat{y}_{i,t-1}, s_{i,t},c_{i,t-1})}Q_{\Psi}(s_{i,t},y_{i,t})
\end{array}
\label{eq:qobjective}
\end{equation}
Similar to the value network training, the $Q$-function estimation is a regression problem and the Mean Squared Error (MSE) is used for training it.
However, one of the differences between the $Q$-function training and value function training is the way in which the true estimates are chosen. In value function estimation, the ground-truth data is used to calculate the true reward-to-go as $v_i=\sum_{t^{\prime}=t}^{T}r(s_{i,t^{\prime}},y_{i,t^{\prime}})$, however, in $Q$-learning, the estimation from the network approximator itself is used to train the regression model:
\begin{equation}
\begin{array}{c}
\mathcal{L}(\Psi) = \frac{1}{2}\sum_i || Q_{\Psi}(s_i,y_i)-q_i ||^2\\
q_i = r_t + \gamma max_{y^{\prime}} Q_{\Psi}(s^{\prime}_i,y^{\prime}_i)
\end{array}
\label{eq:qloss}
\end{equation}
where $s^{\prime}_i$ and $y^{\prime}_i$ are the state and action at the next time, respectively.
Although the $Q$-value estimation has no direct relation to the true $Q$-values calculated using ground-truth data, in practice, it is known to provide good estimation and faster training due to not collecting ground-truth reward at each step of the training.
However, there are no rigorous studies that analyze how far are these estimates from the true $Q$-values.
As shown in Eq. (\ref{eq:qloss}), the true $Q$ estimations is calculated using the estimation from network approximator at time $t+1$, i.e. $max_y^{\prime} Q_{\Psi}(s^{\prime}_i,y^{\prime}_i)$.
Although, not relying on the true ground-truth estimation and explicitly using the reward function might seems to be a bad idea, however in practice it is shown that these models provide better and more robust estimators. Therefore, the training process in $Q$-learning consists of first collecting a dataset of experiences $e_t = (s_t, y_t, s_{t^{\prime}}, r_t)$ during training our Actor model and then use them to train the network approximator. This is the standard way of training the $Q$-network and was frequently used in earlier temporal-difference learning models. But, there is a problem with this method.
Generally, the Actor-Critic models with neural network as function estimator are tricky to train and unless there are guarantees that the estimator is good, the model does not converge.
Although the original $Q$-learning method is proven to converge~\cite{watkins1992q,tsitsiklis1994asynchronous}, when a neural networks is used to approximate the estimator, the convergence guarantee no longer holds.
Usually, since samples are coming from a specific sets of sequences, there is a correlation between the samples that are chosen to train the model.
Thus, this may cause any small updates to Q-network to significantly change the data distribution, and ultimately affects the correlations between $Q$ and the target values.
Recently, Mnih \textit{et al.}~\cite{mnih2015human} proposed using an \textit{experience buffer}~\cite{lin1992self}\footnote{In some literatures, it is called a \textit{replay buffer}} to store the experiences from different sequences and then randomly select a batch from this dataset and train the $Q$-network.
Similar to the off-policy AC model, one benefit of using this buffer is the potential to increase  efficiency of the model by re-using the experiences in multiple updates and reducing the variance of the model.
Since, by sampling uniformly from the buffer, the correlation of samples used in the updates is reduced.
As another improvement to the \textit{experience buffer}, a prioritized version of this buffer is used in which, to select the mini-batches during training, only samples that have low temporal difference error are selected~\cite{schaul2015prioritized}.
Algorithm~\ref{alg:qlearning} provides the pseudo-code for a $Q$-learning algorithm called Deep $Q$-Network or DQN.

\begin{algorithm}[t]
\begin{algorithmic}
\footnotesize
\State \textbf{Input}: Input sequences ($X$), ground-truth output sequences ($Y$),
\State and preferably a pre-trained Actor model ($\pi_\theta$).
\State \textbf{Output}: Trained Actor and Critic models.
\State \textbf{Training Steps}:
\State Initialize the Actor (Seq2seq) model, $\pi_\theta$.
\State Initialize the Critic ($Q$-Net) model, $Q_{\Psi}$.
\While{not converged}
    \State \textbf{Training Seq2seq Model}:
    \State Select a batch of size $N$ from $X$ and $Y$.
    \State Sample $N$ full sequences of actions based on the Actor
    \State model, $\pi_\theta$.
    \For{$n=1,\cdots,N$}
        \For{$t=1,\cdots,T$}
            \State Collect experience $e_t = (s_t, y_t, s_{t^{\prime}}, r_t)$ and
            \State add them to the \textit{experience buffer}.
        \EndFor
    \EndFor
    \State
    \State \textbf{Training $Q$-Net}:
    \State Select a batch of size $N_q$ from the \textit{experience buffer}.
    \State based on the reward.
    \For{$n=1,\cdots,N_q$}
        \State Estimate $\hat{q_n} = Q_{\Psi}(s_n,y_n)$.
        \State Calculate the true estimation:
        \State $q_n=\left\{
        		\begin{tabular}{ll}
		$r_n$ & $s^{\prime}_n$==EOS\\
		$r_n + \gamma max_{y^{\prime}}Q_{\Psi}(s^{\prime}_n,y^{\prime}_n)$ & otherwise.
		\end{tabular}
	\right.$
	\State Store $(\hat{q}_n, q_n)$.
    \EndFor
    \State \textbf{Updating $Q$-Net}:
    \State Minimize the loss using Eq. (\ref{eq:qloss}).
    \State Update the parameters of network, $\Psi$.
    \State
    \State \textbf{Updating Seq2seq Model}:
    \State Use the estimated $Q$ values for $\hat{q}_n = Q_{\Psi}(s_n,y_n)$.
    \State to calculate the loss using Eq. (\ref{eq:qobjective}).
    \State Update parameters of the model using $\theta \leftarrow \theta + \alpha \nabla_{\theta} \mathcal{L}(\theta)$.
\EndWhile
\end{algorithmic}
\caption{Deep $Q$-Learning}
\label{alg:qlearning}
\end{algorithm}

\subsection{Advanced Q-Learning}
\label{section:advq}
\subsubsection{Double Q-Learning}
One of the problems with the Deep $Q$-Network (DQN) is the overestimation of $Q$-values as discussed in~\cite{hasselt2010double,thrun1993issues}.
Specifically, the problem lies in the fact that the ground-truth reward is not used to train these models and the same network is used to calculate both the estimation of network $Q_{\Psi}(s_i,y_i)$ and true values for regression training, $q_i$.
To alleviate this problem, one could use two different networks in which the first one chooses the best action when calculating $max_{y^{\prime}}Q_{\Psi}(s^{\prime}_n,y^{\prime}_n)$ and the other calculate the estimation of $Q$ value, i.e., $Q_{\Psi}(s_i,y_i)$.
In practice, a modified version of the current DQN network is used as the second network in which the current network freezes its parameters for a certain period of time and updates the second network, periodically.
Let us call the second network as the target network with parameters $\Psi^{\prime}$.
We know that $max_{y^{\prime}}Q_{\Psi}(s^{\prime}_n,y^{\prime}_n)$ is the same as choosing the best action according to the network $Q_{\Psi}$.
Therefore, this equation can be re-written as $Q_{\Psi}(s^{\prime}_t,\argmax_{y^{\prime}_t}Q_{\Psi}(s^{\prime}_t,y^{\prime}_t))$.
As shown in this equation, $Q_{\Psi}$ is used for both calculating the $Q$-value and finding the best action.
Given a target network, the best action is chosen using our target network and the $Q$-value is estimated using the current network.
Hence, using the target network, $Q_{\Psi^{\prime}}$, the $Q$-estimation will be given as follows:
\begin{equation}
q_t=\left\{
        	\begin{tabular}{ll}
	$r_t$ & $s^{\prime}_n$==EOS\\
	$r_t + \gamma Q_{\Psi}(s^{\prime}_t,\argmax_{y^{\prime}_t}Q_{\Psi^{\prime}}(s^{\prime}_t,y^{\prime}_t))$ & otherwise.
	\end{tabular}
\right.
\label{eq:ddqn}
\end{equation}
where EOS stands for the End-Of-Sequence action.
This method is called Double DQN (DDQN)~\cite{hasselt2010double,van2016deep} and is shown to resolve the problem of overestimation in DQN and provides more realistic estimations. However, even this model suffers from the fact that there is no relation between the true $Q$-values and the estimation provided by the network. Algorithm~\ref{alg:ddqn} shows the pseudocode for this model.

\begin{algorithm}[t]
\begin{algorithmic}
\footnotesize
\State \textbf{Input}: Input sequences ($X$), ground-truth output sequences ($Y$),
\State and preferably a pre-trained Actor model ($\pi_\theta$).
\State \textbf{Output}: Trained Actor and Critic models.
\State \textbf{Training Steps}:
\State Initialize the Actor (Seq2seq) model, $\pi_\theta$.
\State Initialize the two Critic models:
\State current $Q$-Net, $Q_{\Psi}$, and target $Q$-net, $Q_{\Psi^{\prime}}$: $Q_{\Psi^{\prime}}\leftarrow Q_{\Psi}$.
\While{not converged}
    \State \textbf{Training Seq2seq Model}:
    \State Select a batch of size $N$ from $X$ and $Y$.
    \State Sample $N$ full sequences of actions based on the Actor
    \State model, $\pi_\theta$.
    \For{$n=1,\cdots,N$}
        \For{$t=1,\cdots,T$}
            \State Collect experience $e_t = (s_t, y_t, s_{t^{\prime}}, r_t)$ and
            \State add them to the \textit{experience buffer}.
        \EndFor
    \EndFor
    \State
    \State \textbf{Training $Q$-Net}:
    \State Select a batch of size $N_q$ from the \textit{experience buffer}
    \State based on the reward.
    \For{$n=1,\cdots,N_q$}
        \State Estimate $\hat{q_n} = Q_{\Psi}(s_n,y_n)$
        \State Calculate the true estimation:
        \State $q_n=\left\{
        		\begin{tabular}{ll}
		$r_n$ & $s^{\prime}_n$==EOS\\
		$r_n + \gamma Q_{\Psi}(s^{\prime}_n,\argmax_{y^{\prime}_t}Q_{\Psi^{\prime}}(s^{\prime}_t,y^{\prime}_t))$ & otherwise.
		\end{tabular}
	\right.$
	\State Store $(\hat{q}_n, q_n)$.
    \EndFor
    \State \textbf{Updating current $Q$-Net}:
    \State Minimize the loss using Eq. (\ref{eq:qloss}).
    \State Update the parameters of network, $\Psi$.
    \State
    \State \textbf{Updating target $Q$-Net every $N_u$ iterations}:
    \State $\Psi^{\prime} \leftarrow \Psi$ or using Polyak averaging:
    \State $\Psi^{\prime} \leftarrow \tau \Psi^{\prime} + (1-\tau) \Psi$, $\tau=\frac{1000-(\text{Current Step} \% 1000)}{1000}$.
    \State
    \State \textbf{Updating Seq2seq Model}:
    \State Use the estimated $Q$-values for $\hat{q}_n = Q_{\Psi}(s_n,y_n)$
    \State to calculate the loss using Eq. (\ref{eq:qobjective}).
    \State Update parameters of the model using $\theta \leftarrow \theta + \alpha \nabla_{\theta} \mathcal{L}(\theta)$.
\EndWhile
\end{algorithmic}
\caption{Double Deep $Q$-Learning}
\label{alg:ddqn}
\end{algorithm}

\subsubsection{Dueling Networks}
\begin{figure*}
\centering
\includegraphics[width=0.9\textwidth]{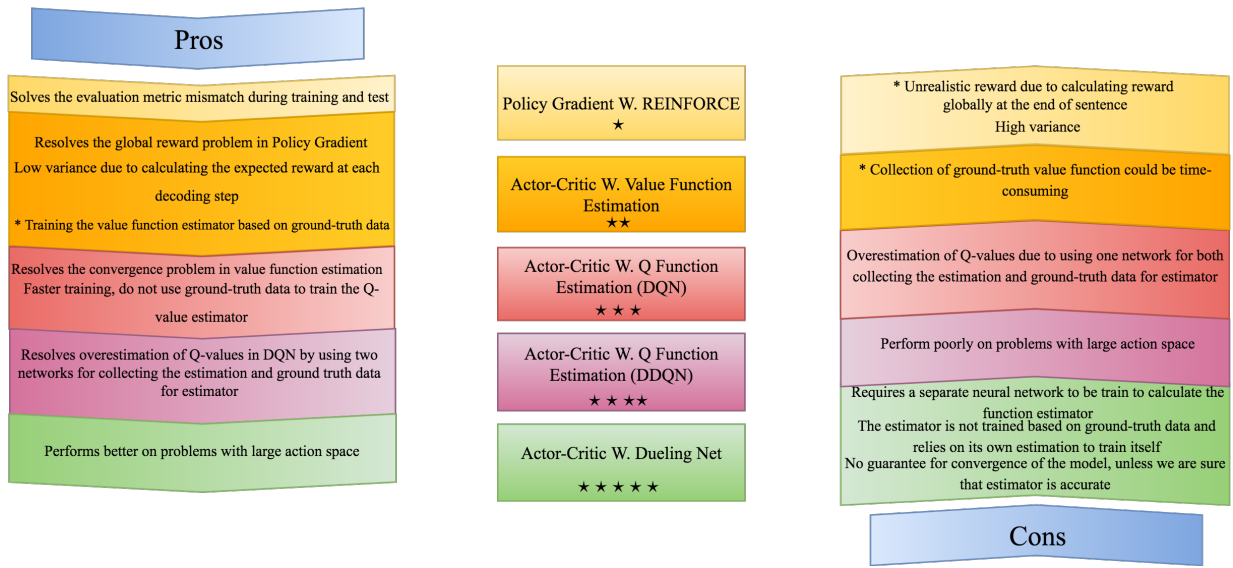}
\caption{A list of advantages and drawbacks of different RL models.
The advantages are listed such that each method covers all the strengths of its previous methods and drawbacks are listed such that each method have all the weaknesses of the previous ones.
For instance, Actor-Critic w. Dueling Net have all the pros of the previous models listed above it and Actor-Critic w. Value Function Estimation suffers from all the cons of the methods listed below it.
The features that are also model-dependent are shown with `$*$' and those features do not exist in any other model.
Each `$\star$' shows how hard it is to implement these models in a real-world application.}
\label{fig:proscons}
\end{figure*}

DDQN tried to solve one of the problems with DQN model by using two networks in which the target network selects the next best action while the current network estimates the $Q$-values given the action selected by the target.
However, in most applications, it is unnecessary to estimate the value of each action choice.
This is especially important in discrete problems with a large sets of possible actions where only a small portion of actions are suitable. For instance, in text summarization the output of the model is a vector of the distribution over the vocabulary and therefore, the output has the same dimension as the vocabulary size which is usually selected to be between 50K to 150K. In most of the applications that use DDQN, the action space is limited to less than a few hundred. For instance, in an Atari game, the possible actions could be to move left, right, up, down, and shoot. Therefore, using DDQN would be easy for these types of applications. Recently, Wang \textit{et al.}~\cite{wang2015dueling} proposed the idea of using a dueling net to overcome this problem. In their proposed method, rather than estimating the $Q$-values directly from the $Q$-net, two different values are estimated for the value function and advantage function as follows:
\begin{equation}
Q_\Psi(s_t, y_t) = V_\Psi(s_t) + A_\Psi(s_t,y_t)
\label{eq:duelingnetp}
\end{equation}
In order to be able to calculate the $V_\Psi(s_t)$, the value estimates is replicated $|\mathcal{A}|$ times.
However, as discussed in~\cite{wang2015dueling}, using Eq. (\ref{eq:duelingnet}) to calculate the $Q$ is bad and can potentially yield poor performance since Eq. (\ref{eq:duelingnet}) is unidentifiable in the sense that a constant can be added to $V_\Psi(s_t)$ and subtracted the same constant from $A_\Psi(s_t,y_t)$.
To solve this problem, the authors suggested to force the advantage estimator to have a zero at the selected action:
\begin{equation}
Q_\Psi(s_t, y_t) = V_\Psi(s_t) + \Big(A_\Psi(s_t,y_t)-\max_{y}A_\Psi(s_t,y)\Big)
\label{eq:duelingmax}
\end{equation}
This way, for the action $y^{*}=\argmax_{y}Q_{\Psi}(s_t,y)=\argmax_{y}A_{\Psi}(s_t,y)$, $Q_{\Psi}(s_t,y^{*})=V_{\Psi}(s_t)$ is obtained.
As an alternative to Eq. (\ref{eq:duelingmax}) and to make the model more stable, the author suggested to replace the max operator with average:
\begin{equation}
Q_\Psi(s_t, y_t) = V_\Psi(s_t) + \Big(A_\Psi(s_t,y_t)-\frac{1}{|\mathcal{A}|}\sum_yA_\Psi(s_t,y)\Big)
\label{eq:duelingnet}
\end{equation}
Note that the dueling net will not decrease the number of actions but will provide a better normalization over the target distribution.
Similar to DQN and DDQN, this model also suffers from the fact that there is no relation between the true values of $Q$-function and the estimation provided by the network.
In Section~\ref{rlseq}, we propose a simple and effective solution to overcome this problem by doing schedule sampling between the $Q$-value estimations and true $Q$-values to pre-train our function approximator.
Fig.~\ref{fig:proscons} summarizes some of the strengths and weaknesses of these different RL methods.

\section{Combining RL with Seq2seq Models}
\label{rlseq2seq}
In this section, we will provide some of the recent models that combined the seq2seq training with reinforcement learning techniques. For most of these models, the main goal is to solve the train/test evaluation mismatch problem which exists in all previously described seq2seq models. This is usually done by adding a reward function to the training objective. There are a growing number of research works that used the REINFORCE algorithm to improve the current state-of-the-art seq2seq models. However, more advanced techniques such as Actor-Critic models, DQN and DDQN, have not been used often for these tasks. As mentioned earlier, one the main difficulties of using $Q$-Learning and its derivatives is the large action space for seq2seq models. For instance, in text summarization, the model should provide estimates for each word in the vocabulary and therefore the estimation could be inferior even with a well trained model. Due to these reasons, researchers mostly focused on the easier yet problematic approaches such as REINFORCE algorithm to train the seq2seq model. Therefore, combining the power of $Q$-Learning training with seq2seq model is still considered to be an open area of research. Table~\ref{table:seq2seqrl} shows the policy, action, and reward function for each seq2seq task and Table~\ref{table:seq2seqmodel} summarizes these models along with the respective seq2seq application and specific RL algorithm they used to improve that application.

\begin{table*}
\centering
\caption{Policy, Action, and Reward function for different seq2seq tasks.}
\label{table:seq2seqrl}
\begin{tabular}{|l|l|l|l|}
\hline
\multicolumn{1}{|c|}{\textbf{Seq2seq Task}}                                                                                  & \multicolumn{1}{c|}{\textbf{Policy}}                                                       & \multicolumn{1}{c|}{\textbf{Action}}                                                                                                                           & \multicolumn{1}{c|}{\textbf{Reward}}                                                      \\ \hline
\begin{tabular}[c]{@{}l@{}}Text Summarization\\ Headline Generation\\ Machine Translation\\ Question Generation\end{tabular} & \begin{tabular}[c]{@{}l@{}}Attention-based models, \\ pointer-generators, etc.\end{tabular} & \begin{tabular}[c]{@{}l@{}}Selecting the next token \\ for summary, headline, \\ and translation\end{tabular}                                                  & ROUGE, BLEU                                                                               \\ \hline
Question Answering                                                                                                           & Seq2seq model                                                                              & \begin{tabular}[c]{@{}l@{}}Selecting the answer from \\ a vocabulary or selecting the\\ start and end index of the\\ answer in the input document\end{tabular} & F1 Score                                                                                  \\ \hline
\begin{tabular}[c]{@{}l@{}}Image Captioning\\ Video Captioning\end{tabular}                                                   & seq2seq model                                                                              & \begin{tabular}[c]{@{}l@{}}Selecting the next token for\\ the caption\end{tabular}                                                                             & CIDEr, SPICE, METEOR                                                                             \\ \hline
Speech Recognition                                                                                                           & Seq2seq model                                                                              & \begin{tabular}[c]{@{}l@{}}Selecting the next token for\\ the speech\end{tabular}                                                                              & \begin{tabular}[c]{@{}l@{}}Connectionist Temporal\\  Classification (CTC)\end{tabular}    \\ \hline
Dialog Generation                                                                                                            & Seq2seq model                                                                              & Dialogue utterance to generate                                                                                                                                 & \begin{tabular}[c]{@{}l@{}}BLEU\\ Length of dialogue\\ Diversity of dialogue\end{tabular} \\ \hline
\end{tabular}
\end{table*}

\subsection{Policy Gradient and REINFORCE Algorithm}
As mentioned in Section~\ref{section:pg}, in Policy Gradient (PG), the reward of the sampled sequence is observed at the end of the sequence generation and back-propagate that error equally to all the decoding steps according to Eq. (\ref{eq:oloss}).
Also, we talked about the exposure bias problem that exists in seq2seq models during training the decoder because of using the Cross-Entropy (CE) error.
The idea of improving generation by letting the model use its own predictions at training time was first proposed by Daume III \textit{et al.}~\cite{daume2009search}. Based on their proposed method, SEARN, the structured prediction problems can be cast as a particular instance of reinforcement learning. The basic idea is to let the model use its own predictions at training time to produce a sequence of actions (e.g., the choice of the next word). Then, a greedy search algorithm is run to determine the optimal action at each time step, and the policy is trained to predict that action. An imitation learning framework was proposed by Ross \textit{et al.}~\cite{ross2011reduction} in a method called DAGGER, where an oracle of the target word given the current predicted word is required. However, for tasks such as text summarization, computing the oracle is infeasible due to the large action space. This problem was later addressed by the `\textit{Data As Demonstrator (DAD)}' model~\cite{venkatraman2015improving} where the target action at step $k$ is the $k^{th}$ action taken by the optimal policy. One drawback of DAD is that at every time step the target label is always selected from the ground-truth data and if the generated summaries are shorter than the ground-truth summaries, the model still forces to generate outputs that could already exist in the model. One way to avoid this problem in DAD is to use a method called End2EndBackProp~\cite{ranzato2015sequence} in which, at each step $t$, the top-$k$ actions are retrieved from the model and the normalized probabilities of these actions are used as weights (of importance) and the normalized combination of their representation is fed to the next decoding step.

%

Finally, REINFORCE algorithm~\cite{williams1992simple} tries to overcome all these problems by using the PG rewarding function and avoiding the CE loss by using the sampled sequence as the ground-truth to train the seq2seq model, Eq. (\ref{eq:olossN}).
In real-world applications, the training is usually started with the CE loss and a pre-trained model is acquired.
Then, the REINFORCE algorithm is used to train the model.
Some of the earliest adoptions of REINFORCE algorithm for training seq2seq models are in computer vision~\cite{mnih2014recurrent,ba2014multiple}, image captioning~\cite{xu2015show}, and speech recognition~\cite{graves2014towards}. Recently, other researchers showed that using a combination of CE loss and REINFORCE loss could yield a better result than just simply performing the pre-training. In these models, the training is started by using the CE loss and is slowly switched from CE loss to REINFORCE loss to train the model.
There are various ways in which one can do the transition from CE loss to REINFORCE loss. Ranzato \textit{et al.}~\cite{ranzato2015sequence} used an incremental scheduling algorithm called `MIXER' which combines DAGGER~\cite{ross2011reduction} with DAD~\cite{venkatraman2015improving} methods. In this method, the RNN is trained with the cross-entropy loss for $N_{CE}$ epochs using the ground-truth sequences. This ensures that the model starts off with a much better policy than random because now the model can focus on promising regions of the search space. Then, they use an annealing schedule in order to gradually teach the model to produce stable sequences. Therefore, after the initial $N_{CE}$ epochs, they continue training the model for $N_{CE}+N_R$ epochs, such that, for every sequence, they use the $\mathcal{L}_{CE}$ for the first $(T - \delta)$ steps, and the REINFORCE algorithm for the remaining $\delta$ steps. The MIXER model was successfully used on a variety of tasks such as text summarization, image captioning, and machine translation.

Another way to handle the transition from using CE loss to REINFORCE loss is to use the following combined loss:
\begin{equation}
\mathcal{L}_{mixed} = \eta \mathcal{L}_{REINFORCE} + (1-\eta) \mathcal{L}_{CE}
\label{eq:mixedloss}
\end{equation}
where $\eta\in(0,1)$ is the parameter that controls the transition from CE to REINFORCE loss.
In the beginning of the training, $\eta=0$ and the model completely relies on CE loss, while as the training progresses, the $\eta$ value is increased in order to slowly reduce the effect of CE loss.
By the end of the training process (where $\eta=1$), the model completely uses the REINFORCE loss for training.
This mixed training loss was used in many of the recent works on text summarization~\cite{paulus2017deep,li2018actor,wang2018reinforced,narayan2018ranking}, paraphrase generation~\cite{li2017paraphrase}, image captioning~\cite{rennie2016self}, video captioning~\cite{pasunuru2017reinforced}, speech recognition~\cite{zhou2017improving}, dialogue generation~\cite{li2016deep}, question answering~\cite{hu2017reinforced}, and question generation~\cite{yuan2017machine}.

\subsection{Actor-Critic Models}
One of the problems with the PG model is that we need to sample the full sequences of actions and observe the reward at the end of the generation. This, in general, will be problematic since the error of generation accumulates over time and usually for long sequences of actions, the final sequence is so far away from the ground-truth sequence. Thus, the reward of the final sequence would be small and model would take a lot of time to converge. To avoid this problem, Actor-Critic models observe the reward at each decoding step using the Critic model and fix the sequence of future actions that the Actor is going to take. The Critic model usually tries to maximize the advantage function through the estimation of value function or $Q$-function. As one of the early attempts of using AC models, Bahdanau \textit{et al.}~\cite{bahdanau2016actor} and He \textit{et al.}~\cite{he2017decoding} used this model for the problem of machine translation. In~\cite{bahdanau2016actor}, the author used temporal-difference (TD) learning for advantage function estimation by considering the $Q$-value for the next action, i.e., $Q(s_t, y_{t+1})$, as a surrogate for the its true value at time $t$, i.e., $V_{\Psi}(s_t)$. We mentioned that for a deterministic policy, $y^{*}=\argmax_yQ(s,y)$, it follows that $Q(s,y^{*})=V(s)$. Therefore, the $Q$-value for the next action could be used as the true estimates of the value function at current time.
To accommodate for the large action space, they also use the shrinking estimation trick that was used in dueling net to push the estimate to be closer to their means. Additionally, the Critic training is done through the following mixed objective function:
\begin{equation}
\begin{array}{c}
\mathcal{L}(\Psi) = \frac{1}{2}\sum_i || Q_{\Psi}(s_i,y_i)-q_i ||^2 + \eta \bar{Q}_i \\
\bar{Q}_i = \sum_y \Big(Q_{\Psi}(y,s_i)-\frac{1}{|\mathcal{A}|}\sum_{y^{\prime}} Q_{\Psi}(y^{\prime},s_i)\Big)
\end{array}
\label{eq:aclossmt}
\end{equation}
where $q_i$ is the true estimation of $Q$ from a delayed Actor.
The idea of using delayed Actor is similar to the idea used in Double $Q$-Learning where a delayed target network is used to get the estimation of the best action.
Later, Zhang \textit{et al.}~\cite{zhang2017actor} used a similar model on image captioning task.

He \textit{et al.}~\cite{he2017decoding} proposed a value network that uses a semantic matching and a context-coverage module and passed them through a dense layer to estimate the value function. However, their model requires a fully-trained seq2seq model to train the value network. Once the value network is trained, they use the trained seq2seq model and trained value estimation model to do the beam search during translation. Therefore, the value network is not used during the training of the seq2seq model. During inference, however, similar to the AlphaGo model~\cite{silver2016mastering}, rather multiplying the advantage estimates (value or $Q$ estimates) to the policy probabilities (like in Eq. (\ref{eq:acinf})), they combine the output of the seq2seq model and the value network as follows:
\begin{equation}
\eta \times \frac{1}{T}\log{\pi(\hat{y}_{1\cdots T}|X)} + (1-\eta)\times \log{V_{\Psi}(\hat{y}_{1\cdots T})}
\end{equation}
where $V_{\Psi}(\hat{y}_{1\cdots T})$ is the output of the value network and $\eta$ controls the effect of each score.

In a different model, Li \textit{et al.}~\cite{li2017learning} proposed a model that controls the length of seq2seq model using RL-based ideas. They train a $Q$-value function approximator which estimates the future outcome of taking an action $y_t$ in the present and then incorporate it into a score $S(y_t)$ at each decoding step as follows:
\begin{equation}
S(y_t) = \log{\pi(y_t|y_{t-1},s_t)+\eta Q(X, y_{1\cdots t})}
\end{equation}
Specifically, the $Q$ function, in this work, takes only the hidden state at time $t$ and estimates the length of the remaining sequence. While decoding, they suggest an inference method that controls the length of the generated sequence as follows:
\begin{equation}
\hat{y}_t = \argmax_{y}\log{\pi(y|\hat{y}_{1\cdots t-1},X)}-\eta ||(T-t)- Q_{\Psi}(s_t)||^2
\end{equation}

Recently, Li \textit{et al.}~\cite{li2018actor} proposed an AC model which uses a binary classifier as the Critic. In this specific model, the Critic tries to distinguish between the generated summary and the human-written summary via a neural network binary classifier. Once they pre-trained the Actor using CE loss, they start training the AC model alternatively using PG and the classifier score is considered as a surrogate for the value function. AC and PG were used also in the work of Liu \textit{et al.}~\cite{liu2017improved} where they combined AC and PG learning along with importance sampling to train a seq2seq model for image captioning. In this method, they used two different neural networks for $Q$-function estimation, i.e., $Q_{\Psi}$, and value estimation, i.e., $V_{\Psi^{\prime}}$. They also used a mixed reward function that combines a weighted sum of $ROUGE$, $BLEU$, $METEOR$, and $CIDEr$ measures to achieve a higher performance on this task.

\begin{table*}
\centering
\caption{A summary of seq2seq applications that used various RL methods.}
\label{table:seq2seqmodel}
\begin{tabular}{|c|c|c|c|c|c|}
\hline
\multicolumn{1}{|c|}{\textbf{Reference}} & \multicolumn{1}{c|}{\textbf{\begin{tabular}[c]{@{}c@{}}Suffers From\\ Exposure Bias\end{tabular}}} & \multicolumn{1}{c|}{\textbf{\begin{tabular}[c]{@{}c@{}}Mismatch on\\ Train/Test Measure\end{tabular}}} & \multicolumn{1}{c|}{\textbf{\begin{tabular}[c]{@{}c@{}}Observe Full\\ Reward\end{tabular}}} & \multicolumn{1}{c|}{\textbf{\begin{tabular}[c]{@{}c@{}}RL\\ Algorithm\end{tabular}}}    & \multicolumn{1}{c|}{\textbf{\begin{tabular}[c]{@{}c@{}}Seq2seq\\ Application\end{tabular}}}                                                                                 \\ \hline
\multicolumn{6}{|c|}{\textbf{Policy Gradient Based Models}}                                                                                                                                                                                 \\ \hline
SEARN~\cite{daume2009search}                   & No                & Yes                     & No Reward           & PG              & \begin{tabular}[c]{@{}c@{}}Sequence Labeling\\ Syntactic Chunking\end{tabular}                      \\ \hline
DAD~\cite{venkatraman2015improving}            & No                & Yes                     & No Reward           & PG              & Time-Series Modeling                                                                                \\ \hline
Qin \textit{et al.}~\cite{qin2018robust}            & No                & Yes                     & Yes           & PG              & Relation Extraction                                                                                \\ \hline
Yin \textit{et al.}~\cite{yin2018deep}            & No                & Yes                     & Yes           & PG              & Pronoun Resolution                                                                                \\ \hline
MIXER\cite{ranzato2015sequence}               & No                & No                      & Yes                 & PG w. REINFORCE & \begin{tabular}[c]{@{}c@{}}Machine Translation\\ Text Summarization\\ Image Captioning\end{tabular} \\ \hline
Wu \textit{et al.}~\cite{wu2018learning}                   & No                & No                      & Yes                 & PG w. REINFORCE & Text Summarization                                                                                 \\ \hline
Narayan \textit{et al.}~\cite{narayan2018ranking}                   & No                & No                      & Yes                 & PG w. REINFORCE & Text Summarization                                                                                 \\ \hline
Kreutzer \textit{et al.}~\cite{kreutzer2018reliability}                   & No                & No                      & Yes                 & PG w. REINFORCE & \begin{tabular}[c]{@{}c@{}}Machine Translation w.\\ Human Bandit Feedback\end{tabular}                                                                                 \\ \hline
Pan \textit{et al.}~\cite{pan2018discourse}                   & No                & No                      & Yes                 & PG w. REINFORCE & Natural Language Inference                                                                                 \\ \hline
Liang \textit{et al.}~\cite{liang2017neural}                   & No                & No                      & Yes                 & PG w. REINFORCE & Semantic Parsing                                                                                 \\ \hline
Li \textit{et al.}~\cite{li2016deep}                   & No                & No                      & Yes                 & PG w. REINFORCE & Dialogue Generation                                                                                 \\ \hline
Yuan \textit{et al.}~\cite{yuan2017machine}            & No                & No                      & Yes                 & PG w. REINFORCE & Question Generation                                                                                 \\ \hline
Mnih \textit{et al.}~\cite{mnih2014recurrent}          & Yes               & No                      & Yes                 & PG w. REINFORCE & Computer Vision                                                                                     \\ \hline
Ba \textit{et al.}~\cite{ba2014multiple}               & Yes               & No                      & Yes                 & PG w. REINFORCE & Computer Vision                                                                                     \\ \hline
Xu \textit{et al.}~\cite{xu2015show}                   & Yes               & No                      & Yes                 & PG w. REINFORCE & Image Captioning                                                                                    \\ \hline
\multicolumn{6}{|c|}{\textbf{Self-Critic Models with REINFORCE Algorithm}}                                                                                                                                                                  \\ \hline
Rennie \textit{et al.}~\cite{rennie2016self}           & Yes               & No                      & Yes                 & SC w. REINFORCE & Image Captioning                                                                                    \\ \hline
Paulus \textit{et al.}~\cite{paulus2017deep}           & No                & No                      & Yes                 & SC w. REINFORCE & Text Summarization                                                                                  \\ \hline
Wang \textit{et al.}~\cite{wang2018reinforced}           & No                & No                      & Yes                 & SC w. REINFORCE & Text Summarization                                                                                  \\ \hline
Pasunuru \textit{et al.}~\cite{pasunuru2017reinforced} & No                & No                      & Yes                 & SC w. REINFORCE & Video Captioning                                                                                    \\ \hline
Yeung \textit{et al.}~\cite{yeung2016end}              & No                & No                      & Yes                 & SC w. REINFORCE & Action Detection in Video                                \\ \hline
Zhou \textit{et al.}~\cite{zhou2017improving}          & No                & No                      & Yes                 & SC w. REINFORCE & Speech Recognition                                                                                  \\ \hline
Hu \textit{et al.}~\cite{hu2017reinforced}             & No                & No                      & Yes                 & SC w. REINFORCE & Question Answering                                                                                  \\ \hline
\multicolumn{6}{|c|}{\textbf{Actor-Critic Models with Policy Gradient and Q-Learning}}                                                                                                                                                      \\ \hline
He \textit{et al.}~\cite{he2017decoding}               & Yes               & No                      & No                  & AC              & Machine Translation                                                                                 \\ \hline
Li \textit{et al.}~\cite{li2017learning}               & Yes               & No                      & No                  & AC              & \begin{tabular}[c]{@{}c@{}}Machine Translation\\ Text Summarization\end{tabular}                    \\ \hline
Bahdanau \textit{et al.}~\cite{bahdanau2016actor}      & Yes               & No                      & No                  & PG w. AC        & Machine Translation                                                                                 \\ \hline
Li \textit{et al.}~\cite{li2018actor}      & Yes               & No                      & No                  & PG w. AC        & Text Summarization                                                                                 \\ \hline
Chen \textit{et al.}~\cite{chen2018fast}      & Yes               & No                      & Yes                  & PG w. AC        & Text Summarization                                                                                 \\ \hline
Zhang \textit{et al.}~\cite{zhang2017actor}            & Yes               & No                      & No                  & PG w. AC        & Image Captioning                                                                                    \\ \hline
Liu \textit{et al.}~\cite{liu2017improved}             & Yes               & No                      & No                  & PG w. AC        & Image Captioning                                                                                    \\ \hline
DARLA~\cite{higgins2017darla}             & Yes               & No                      & No                  & AC        & Domain Adaptation \\ \hline
\end{tabular}
\end{table*}

\subsection{Current RL-Based Model Issues}
Throughout this paper, we discussed about various situations where using RL provides a better solution than traditional methods.
However, utilizing RL methods creates its own training challenges and in most of the cases the improvement received from these models are not significant.
In this section, we will discuss some of the issues that exist in current RL techniques used for seq2seq problems.
As discussed in Section~\ref{section:rl}, sample efficiency and high variance in RL models is one the of the main issues in applying them to seq2seq problems.
Therefore, models such as RAML~\cite{norouzi2016reward} and SPG~\cite{ding2017cold} are proposed to provide a middle ground between the MLE and RL training.
In RAML~\cite{norouzi2016reward}, a reward-aware perturbation is added to MLE while in SPG~\cite{ding2017cold}, the reward distribution is utilized for effective sampling of policy gradient.
Recently, Tan \textit{et al.}~\cite{tan2018connecting} provided a general formulation that connects the MLE and RL training through Entropy Regularized Policy Optimization (ERPO).
However, even these solutions suffer from their own problems.
RAML arguably suffers from the exposure bias while SPG requires a lot of engineering to work on a specific problem and, as shown in Table~\ref{table:seq2seqrl}, that is why REINFORCE-based models such as MIXER~\cite{ranzato2015sequence} are preferred in most of the current seq2seq problems.

Although REINFORCE-based models are simple to implement and provide better results, training these models is time-consuming and the improvement over baselines is usually marginal.
This is why in most of the current works, these models are only used for fine-tuning purposes.

Aside from these issues, there are problems inherent to specific applications that make it hard for researchers to combine RL techniques with current seq2seq models.
For instance, in most of the NLP problems, the output or action space is massive comparing to the size of actions in a robotic or game-playing problems.
This is mostly due to the fact that in applications such as machine translation, text summarization, and image captioning the size of the output is equal the size of the vocabulary used during training.
Now, compare this to an agent that plays a simple Atari game which requires deciding on usually less than 20 actions~\cite{mnih2013playing}.
This will show the severity of this problem and the reward sparsity issue that exist in these applications.

Moreover, most of the current seq2seq models which use RL training, rely on well-defined reward function such as BLEU or ROUGE for providing feedback for the model.
Although these are the standard metrics for evaluating various seq2seq models, relying on them creates a different set of problems.
For instance, in abstractive text summarization, ROUGE and BLEU scores are being used as the standard metric for evaluation of summarization models.
However, a good abstractive summary will definitely have a low ROUGE and BLEU score.
This problem could be further investigated and possibly improved by Inverse Reinforcement Learning (IRL)~\cite{ziebart2008maximum} by forcing the model to learn its own rewarding function.
However, to the best of our knowledge, no work has been done in this area.

Recently, new methods are introduced for game playing using RL algorithm which combine the best performing models in this area and apply some of the best practices used in previous models to achieve state-of-the-art results.
Rainbow~\cite{hessel2017rainbow} and Quantile Nets~\cite{dabney2018implicit} are among such frameworks.
In Rainbow~\cite{hessel2017rainbow}, the authors combine DDQN, prioritized experience buffer, dueling net, multi-step learning (using step-based reward rather than general reward), and distributional RL to achieve state-of-the-art in 57 games in the Atari 2600 framework.
A similar ensembling method could also be useful to be applied for seq2seq tasks but this is also left for future works.
\section{RLSeq2Seq: An Open-Source Library for Training Seq2seq Models with RL Methods}
\label{rlseq}
As part of this comprehensive study, we developed an open-source library which implements various RL techniques for the problem of abstractive text summarization. This library is made available at \textit{\url{www.github.com/yaserkl/RLSeq2Seq/}}. Since experimenting each specific configuration of these models, even requires few days of training on GPUs, we encourage researchers, who use this library to build and enhance their own models, to also share their trained model at this website. In this section, we explain some of the important features of our library. As mentioned before, this library provides modules for abstractive text summarization. The core of our library is based on a model called pointer-generator~\footnote{\url{https://github.com/abisee/pointer-generator}}~\cite{see2017get} which itself is based on Google TextSum model~\footnote{\url{https://github.com/tensorflow/models/tree/master/research/textsum}}. We also provide a similar imitation learning used in training REINFORCE algorithm to train the function approximator. This way, we propose training our DQN (DDQN, Dueling Net) using a schedule sampling in which we start training the model in the beginning based on ground-truth $Q$-values while as we move on with the training process, we completely rely on the function estimator to train the network. This could be seen as a pre-training step for the function approximator. Therefore, the model is guaranteed to start by using better ground-truth data since it is exposed to the true ground-truth values versus the random estimation it receives from the model itself. In summary, our library implements the following features:
\begin{itemize}[leftmargin=*]
\item Adding temporal attention and intra-decoder attention that was proposed in~\cite{paulus2017deep}.
\item Adding scheduled sampling along with its differentiable relaxation proposed in~\cite{goyal2017differentiable} and E2EBackProb~\cite{ranzato2015sequence} for solving \textit{exposure bias} problem.
\item Adding adaptive training of REINFORCE algorithm by minimizing the mixed objective loss in Eq. (\ref{eq:mixedloss}).
\item Providing Self-Critic training by adding the greedy reward as the baseline.
\item Providing Actor-Critic training options for training the model using asynchronous training of Value Network, DQN, DDQN, and Dueling Net.
\item Providing options for scheduled sampling for training of the $Q$-Function in DQN, DDQN, and Dueling Net.
\end{itemize}

\subsection{Experiments}
To test the power of some of the studied models in this paper, we have done a range of various experiments using our open-source library.
As mentioned in Section~\ref{rlseq2seq}, most of the RL-based models play as a fine-tuning technique in seq2seq applications.
Thus, we first pre-train our model for 15 epochs using only cross-entropy loss and then add the RL training for another 10 epochs.
Our experiments follows the same setup as to the pointer-generator model~\cite{see2017get} and we only show the results after activating the coverage mechanism.
We activate the coverage mechanism only for the last epoch and select the best model using the evaluation data.
We use a linear scheduling probability as $\epsilon=step/total steps$ and we use $\epsilon= 1$ after activating the coverage so that the model completely relies on its own output for the rest of training and for E2EBackPropagation model, $K$ is set to 4.
All experiments are done using two NVIDIA P100 GPUs one used for training the model and the other for select the best trained model based on the evaluation data.

\subsubsection{Analysis of the results}
Table~\ref{table:exp} shows the results of our experiments based on ROUGE score on this dataset.
All our ROUGE scores  have  a  95\%  confidence  interval  of  at  most $\pm 0.25$ as reported by the official ROUGE script.
In this table, PG stands for Pointer-Generation and SS stands for Scheduled Sampling.
As shown in this table, both scheduled sampling model and E2E model are superior to the pointer-generator.
We have also used our framework to train the Self-Critic Policy Gradient (SCPG) based model proposed by Paulus \textit{et al.}~\cite{paulus2017deep}.
However, as shown in this table, although the SCPG improves the performance of the pointer-generator model, this improvement is very marginal.
This result is totally in contrast with the result in the original paper~\cite{paulus2017deep} and shows that SCPG, as claimed by the authors, will not greatly improve the performance of the pointer-generator model.
One of the main reason for this difference in the result of our experiment with the Paulus \textit{et al.}~\cite{paulus2017deep} paper is that they use a completely different set of hyperparameters for training their model.
For instance, the input for their encoder is 800 words while in our default setting, for all our experiments, it is set to 400.
Also, the vocabulary size is set to 150K and 50K for input and output while our default is set to 50K for both input and output.
Moreover, the size of hidden layers for encoder and decoder in their work is larger that our default values and they also use a pre-trained GloVe~\cite{pennington2014glove} word-embedding for training their model.
Finally, we are comparing all these policy-gradient based models with an Actor-Critic model proposed by Chen \textit{et al.}~\cite{chen2018fast} which holds the state-of-the-art result in text summarization in CNN/DM dataset.
As shown in Table~\ref{table:exp}, this model is superior to any of the policy-gradient based models according to the ROUGE scores.

\subsubsection{Analysis of the training time}
In general, the pointer-generator framework requires more than 3 days of training for an effective results, while this time will also be expanded after adding the self-critic policy gradient.
On average, each batch of training during MLE training will take 2-3 seconds while once we add the SCPG loss this time will be increased to 5-6 seconds which means the whole training time will be double after activation of the RL loss.
On the other hand, the whole training time for the Actor-Critic model before and after RL activation is only a few hours which shows that not only it has superiority in the ROUGE score results but also the training process converges much faster than the other models.

\begin{table}[t!]
\centering
\caption{Analysis of ROUGE F1-Score after coverage and approximate amount of training time for various RL-based techniques.}
\label{table:exp}
\begin{tabular}{|c|c|c|c|c|}
\hline
\multirow{2}{*}{\textbf{Method}}                                                      & \multicolumn{3}{c|}{\textbf{ROUGE}}     & \multirow{2}{*}{\textbf{Training Time}}         \\ \cline{2-4} 
                                                                                & \textbf{1}     & \textbf{2}     & \textbf{L} &      \\ \hline
PG                             & 38.21          & 16.46 & 34.79 & 3-4 days         \\ \hline
\begin{tabular}[c]{@{}c@{}}PG w. E2E\end{tabular} & 38.24 & 16.48 & 34.97 & 3-4 days        \\ \hline
\begin{tabular}[c]{@{}c@{}}PG w. SS\\Argmax-Sampling\end{tabular}      & 38.29 & 16.51 & 35.02 & 3-4 days \\ \hline
\begin{tabular}[c]{@{}c@{}}PG w. SS\\Argmax-Greedy\end{tabular}      & 38.65 & 16.77 & 35.37 & 3-4 days\\ \hline
\begin{tabular}[c]{@{}c@{}}PG w. SCPG\end{tabular} & 38.77 & 16.98 & 35.32 & 6-7 days         \\ \hline
\begin{tabular}[c]{@{}c@{}}Actor-Critic\\(Q-Learning)~\cite{chen2018fast}\end{tabular} & 40.88 & 17.80 & 38.54 & 3-4 hours \\ \hline
\end{tabular}
\end{table}

\section{Conclusion}
\label{section:conclusion}

In this paper, we provided a general overview of a specific type of deep learning models called sequence-to-sequence (seq2seq) models and discussed some of the recent advances in combining training of these models with Reinforcement Learning (RL) techniques. Seq2seq models are common in a wide range of applications from machine translation to speech recognition. However, traditional models in this field usually suffer from various problems during model training, such as inconsistency between the training objective and testing objective and \textit{exposure bias}. Recently, with advances in deep reinforcement learning, researchers offered various types of solutions to combine the RL training with seq2seq training for alleviating the problems and challenges of training seq2seq models. In this paper, we summarized some of the most important works that tried to combine these two different techniques and provided an open-source library for the problem of abstractive text summarization that shows how one could train a seq2seq model using different RL techniques.

\section*{Acknowledgments}
This work was supported in part by the US National Science Foundation grants IIS-1619028, IIS-1707498 and IIS-1838730.

\bibliographystyle{IEEEtran}
\bibliography{reference}

\begin{IEEEbiography}[{\includegraphics[width=1in,height=1.25in,clip,keepaspectratio]{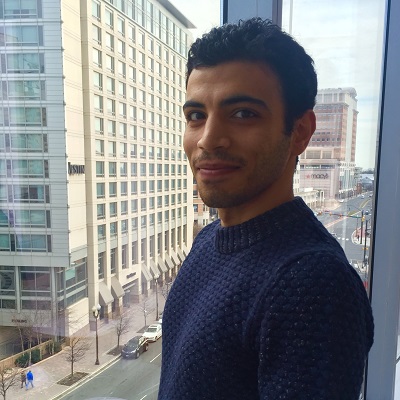}}]{Yaser Keneshloo}
received his Masters degree in Computer Engineering from Iran University of Science and Technology in 2012. Currently, he is a Ph.D candidate in the Department of Computer Science at Virginia Tech. His research interests includes machine learning, data mining, and deep learning.
\end{IEEEbiography}

\begin{IEEEbiography}[{\includegraphics[width=1in,height=1.25in,clip,keepaspectratio]{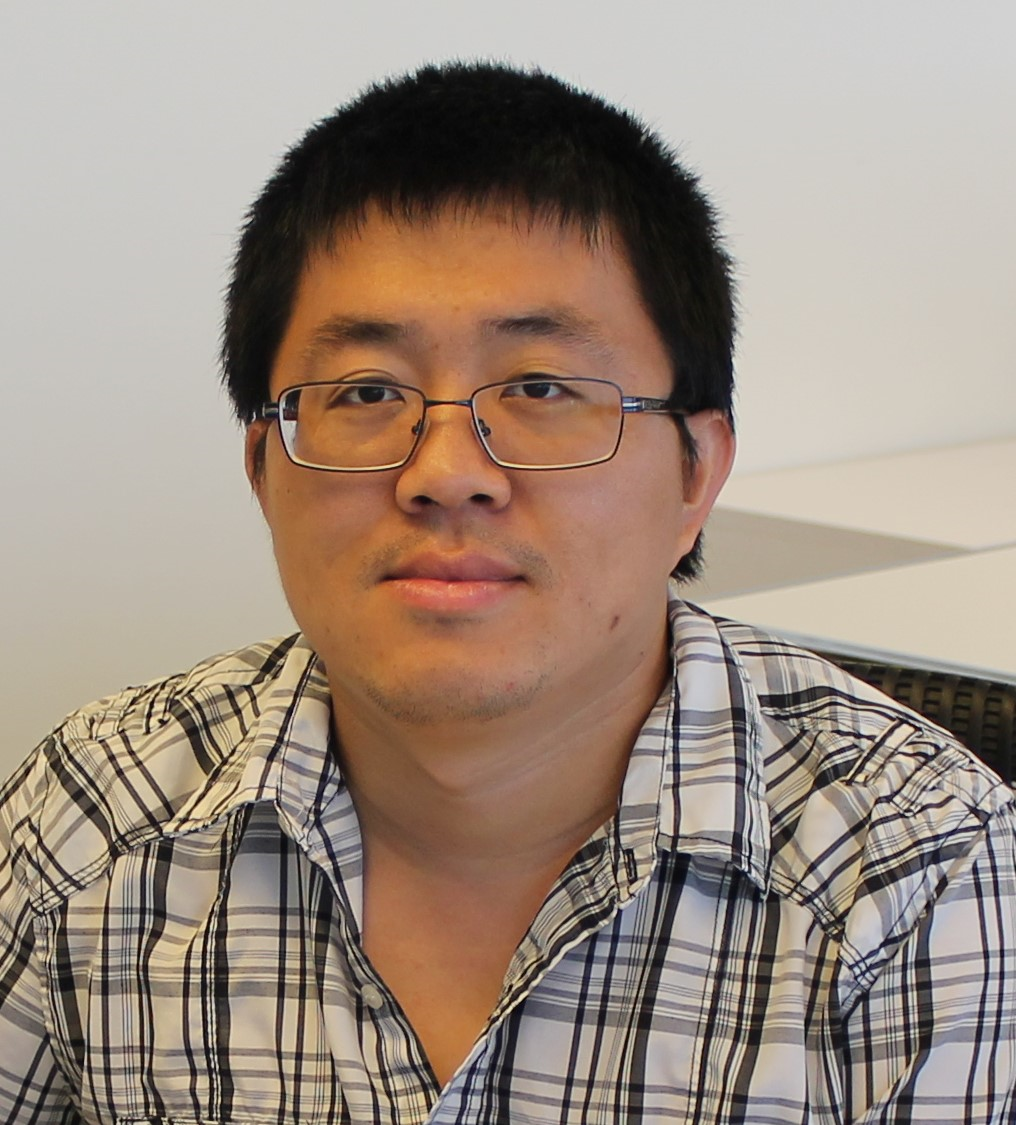}}]{Tian Shi}
Tian Shi received the Ph.D. degree in Physical Chemistry from Wayne State University in 2016. He is working toward the Ph.D. degree in the Department of Computer Science, Virginia Tech. His research interests include data mining, deep learning, topic modeling, and text summarization.
\end{IEEEbiography}

\begin{IEEEbiography}[{\includegraphics[width=1in,height=1.25in,clip,keepaspectratio]{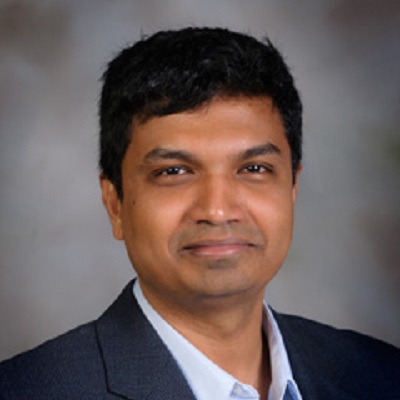}}]{Naren Ramakrishnan}
is the Thomas L. Phillips Professor of Engineering at Virginia Tech. He directs the Discovery Analytics Center, a university-wide effort that brings together researchers from computer science, statistics, mathematics, and electrical and computer engineering to tackle knowledge discovery problems in important areas of national interest. His work has been featured in the Wall Street Journal, Newsweek, Smithsonian Magazine, PBS/NoVA Next, Chronicle of Higher Education, and Popular Science, among other venues. Ramakrishnan serves on the editorial boards of IEEE Computer, ACM Transactions on Knowledge Discovery from Data, Data Mining and Knowledge Discovery, IEEE Transactions on Knowledge and Data Engineering, and other journals. He received his PhD in Computer Sciences from Purdue University.
\end{IEEEbiography}

\begin{IEEEbiography}[{\includegraphics[width=1in,height=1.25in,clip,keepaspectratio]{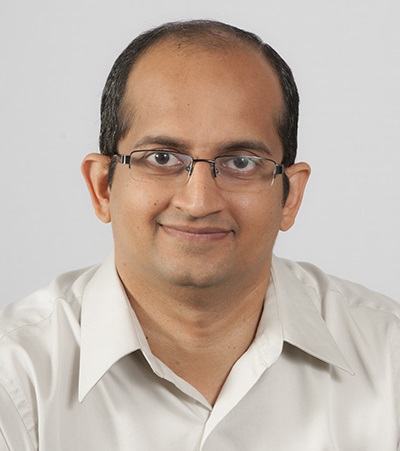}}]{Chandan K. Reddy}
is an Associate Professor in the Department of Computer Science at Virginia Tech. He received his Ph.D. from Cornell University and M.S. from Michigan State University. His primary research interests are Data Mining and Machine Learning with applications to Healthcare Analytics and Social Network Analysis. His research is funded by the National Science Foundation, the National Institutes of Health, the Department of Transportation, and the Susan G. Komen for the Cure Foundation. He has published over 95 peer-reviewed articles in leading conferences and journals. He received several awards for his research work including the Best Application Paper Award at ACM SIGKDD conference in 2010, Best Poster Award at IEEE VAST conference in 2014, Best Student Paper Award at IEEE ICDM conference in 2016, and was a finalist of the INFORMS Franz Edelman Award Competition in 2011. He is an associate editor of the ACM Transactions on Knowledge Discovery and Data Mining and PC Co-Chair of ASONAM 2018. He is a senior member of the IEEE and life member of the ACM.
\end{IEEEbiography}

\end{document}